\documentclass[Numbered,pdflatex,sn-basic]{sn-jnl}

\usepackage{graphicx}%
\usepackage{multirow}%
\usepackage{amsmath,amssymb,amsfonts}%
\usepackage{booktabs}%
\usepackage{array}%
\usepackage{tabularx}%
\usepackage{makecell}%
\usepackage{url}%

\begin{document}

\title[Ability-Residual Decoupled Affective Cognitive Diagnosis]{Ability-Residual Decoupled Modeling for Affective Cognitive Diagnosis}

\author[1]{\fnm{Boyuan} \sur{Zhao}}\email{20243259@snnu.edu.cn}
\author*[1]{\fnm{Meng} \sur{Ye}}\email{yemeng@snnu.edu.cn}

\affil*[1]{\orgdiv{Key Laboratory of Modern Teaching Technology},
\orgname{Shaanxi Normal University},
\orgaddress{\street{199 Chang'an South Road}, \city{Xi'an},
\postcode{710062}, \state{Shaanxi}, \country{China}}}

\abstract{Cognitive diagnosis infers students' concept mastery from response logs. However, students' responses are not determined by mastery alone: non-cognitive factors such as emotion, engagement, and fatigue can also affect performance. Affective cognitive diagnosis therefore extends conventional cognitive diagnosis by incorporating affective states. Existing methods often assume that the cognitive diagnosis backbone has already explained ability, item, and concept effects, so the remaining errors can be attributed mainly to affect. We argue that this assumption can be insufficient in real educational data: item calibration bias, systematic concept bias, personalized student-concept deviations, and latent student-item matching can form stable cognitive residuals. Without an explicit modeling pathway, these residuals may leak into affective representations, producing affect contamination. To address this problem, we propose an ability-residual decoupled framework for affective cognitive diagnosis. The model first captures unmodeled cognitive residuals through student, item, concept, student-concept, and low-rank student-item components, and then uses an affective module to modulate guess/slip effects. A Q-matrix-constrained concept residual attention mechanism adaptively aggregates only item-relevant concept residuals. Experiments on ASSIST2017, ASSIST2012, ASSIST2009, and Junyi with six cognitive diagnosis backbones show response-prediction gains across the reported comparisons and generally improved affect alignment when affect labels are available. Ablation studies, leakage probes, principal component analysis visualization, long-tail analysis, and case studies further indicate that ability residuals absorb stable cognitive bias, reduce cognitive contamination in the affective branch, and enhance the robustness and predictive accuracy of cognitive diagnosis models. Our code is available at \url{https://github.com/psychosiwa/ARACD}.}


\maketitle

\section{Introduction}

Cognitive diagnosis (CD) infers students' mastery of fine-grained knowledge concepts, skills, or cognitive attributes from their responses in exercises, tests, or online learning platforms \cite{ref1}, \cite{ref2}. Unlike evaluation methods that report an overall score or a holistic ability estimate, CD emphasizes interpretable diagnostic results, i.e., which concepts a student has mastered and which remain deficient. Therefore, CD can support personalized exercise recommendation, learning path planning, formative assessment, and adaptive testing in intelligent tutoring systems \cite{ref2,ref3,ref4}. As student response logs continue to accumulate in large-scale online education, CD has emerged as a critical technology connecting educational measurement, student modeling, and intelligent educational decision-making \cite{ref5,ref6,ref7}.

Research on cognitive diagnosis models (CDMs) has evolved from psychometric models to neural diagnostic models. Classic approaches include DINA, IRT, and MIRT. Specifically, DINA uses the Q-matrix to characterize the concepts required by an item and explains response noise through discrete mastery patterns and item parameters, such as guessing and slipping \cite{ref1}, \cite{ref8}. Accordingly, IRT estimates response probabilities from the matching relationship between continuous latent ability and item difficulty \cite{ref9}, while MIRT extends ability to a multidimensional latent space for multi-skill assessment tasks \cite{ref10}. Deep learning has recently been introduced into CD, affording models that can represent more complex interactions among students, items, and cognitive attributes. For instance, NCDM strengthens student-item-concept interaction modeling with neural networks while retaining interpretability \cite{ref5}. Alternative approaches, such as RCD and SCD, improve diagnosis under sparse-data conditions by using relational structures and self-supervised signals, respectively \cite{ref11}, \cite{ref12}. Higher-order ability modeling, global-local cognitive modeling, and attention mechanisms have also been studied to feature complex ability structures, learning dynamics, and item-related concept relations \cite{ref7}, \cite{ref13}.

Although CDMs are designed to explain response behavior through cognitive ability or mastery, an observed response may be driven by latent cognitive factors, latent affective factors, or their joint influence \cite{ref21}. Existing studies report distinct roles for different affective factors, with control-value theory explaining how achievement emotions shape motivation, learning strategies, and performance \cite{ref14}. Empirical work reveals that affective states evolve dynamically during complex learning \cite{ref15}, and that boredom and frustration differ in their incidence, persistence, and consequences across learning environments \cite{ref16}. Furthermore, affect-detection research has introduced methods for estimating learner states \cite{ref17}, whereas affect-aware tutoring studies use these estimates to adapt learner-system interactions and feedback \cite{ref18,ref19,ref20}. Together, these findings lay the path for incorporating affective factors into cognitive diagnosis. Consequently, researchers have introduced affective states into cognitive diagnosis frameworks, leading to a new perspective, affective cognitive diagnosis. These models estimate students' non-cognitive states through an affective module and use them to modulate guess/slip effects, thereby explaining response fluctuations among students with comparable knowledge mastery \cite{ref21}. Notably, when real affect labels are unavailable, these methods can draw on supervised or instance-level contrastive learning and use response labels to construct weak positive and negative relations for learning response-derived affective representations \cite{ref22}, \cite{ref23}.

Existing affective cognitive diagnosis methods implicitly assume that the base cognitive diagnosis backbone has already explained student, item, and concept-related cognitive structure sufficiently, so that the remaining response variation mainly reflects affective states \cite{ref5}, \cite{ref11}, \cite{ref12}, \cite{ref21}. Nevertheless, this assumption can be insufficient in real educational data, as items may contain difficulty calibration bias; concepts may exhibit systematic prediction bias; a student's mastery of a specific concept may deviate from his or her overall ability; and a student-item pair may involve personalized matching effects due to wording familiarity, problem-solving strategies, or hidden skill requirements \cite{ref7}, \cite{ref24}, \cite{ref25}. When these stable cognitive residuals are not explicitly modeled, the affective module may be forced to absorb them. Consequently, the learned affective representation may encode cognitive-residual information associated with items, concepts, and student ability in addition to non-cognitive states. This phenomenon is defined as affect contamination, namely cognitive residual contamination in affective representations. Such contamination weakens interpretability and may force the model to misattribute cognitive residual variation to affective factors, thereby reducing diagnostic reliability \cite{ref17}, \cite{ref21}.

Motivated by this problem, this study develops an ability-residual decoupling framework for affective cognitive diagnosis. It instantiates it as ARACD or ARCACD depending on whether real affect supervision is available. Inspired by bias decomposition and low-rank interaction modeling in recommender systems \cite{ref24,ref25}, the proposed approach uses multiple structured bias terms to express unmodeled cognitive residual variation, including item, student, concept-shared, and student-concept personalized biases. Furthermore, we introduce a low-rank student-item interaction term to capture latent personalized matching effects without parameterizing a complete student-item interaction matrix. Concept-level residuals are aggregated by a Q-matrix-constrained attention mechanism that assigns student--item-pair-conditioned weights to item-related concept-shared and student-concept residuals. The resulting ability residual first corrects the base prediction on the logit scale, after which the affective module performs guess/slip modulation. This ordered design places stable cognitive residual correction before affective modulation, assigning cognitive residual and affective effects to separate modeling pathways \cite{ref5,ref13}. Figure 1 illustrates the affect contamination and ability-residual decoupling: stable cognitive residuals not explained by the backbone may leak into the affective branch, whereas a dedicated ability-residual pathway provides a more appropriate structural destination for this residual information.

\begin{figure}[htbp]
\centering
\includegraphics[width=0.96\linewidth]{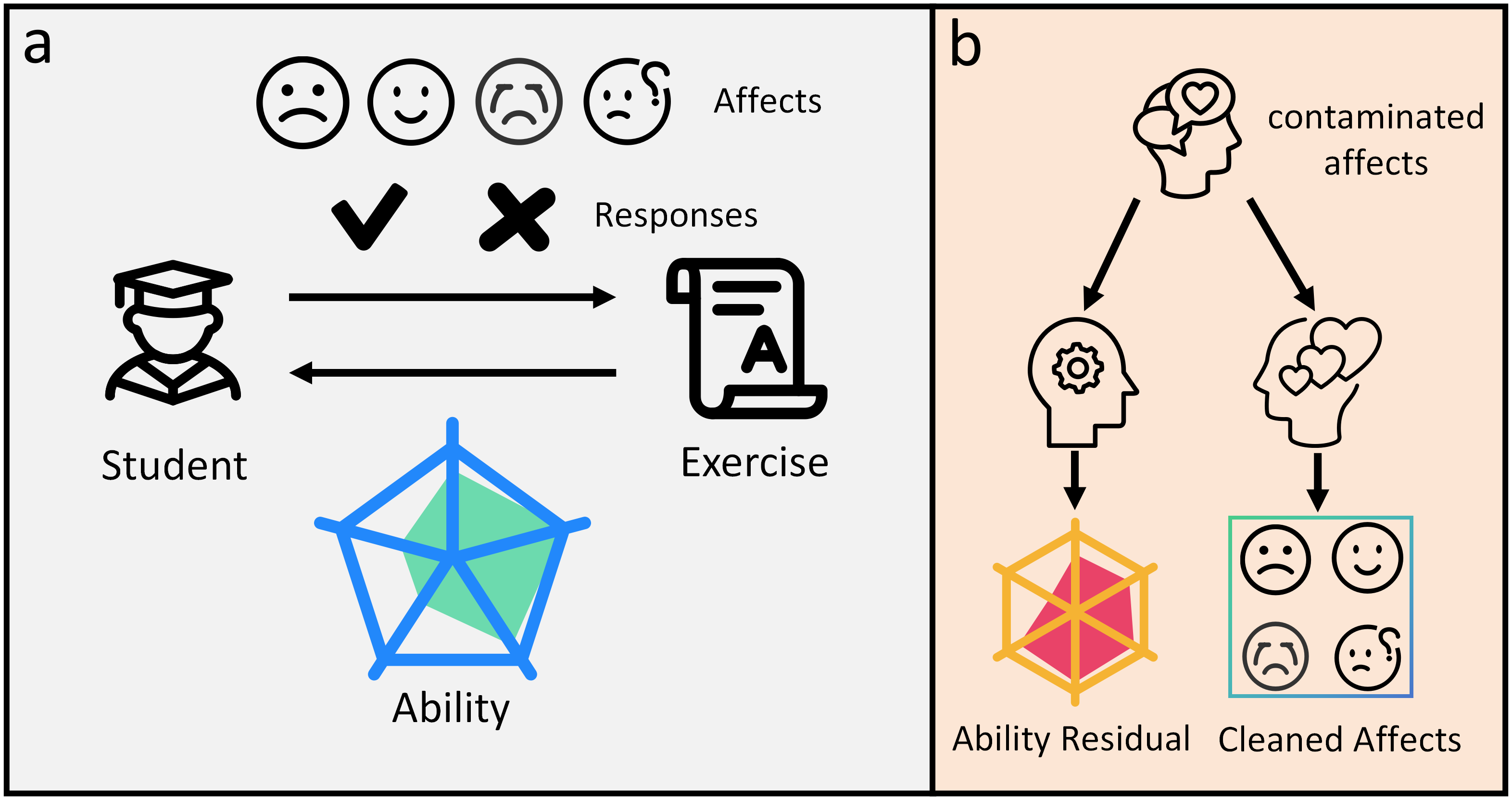}
\caption{Schematic illustration of affect contamination and ability-residual decoupling. Panel (a) illustrates that a student's observed response to an exercise may reflect both cognitive ability and affective state. Panel (b) illustrates the proposed allocation of modeling roles: stable cognitive residual variation is assigned to the ability-residual pathway, reducing its leakage into the affective branch and allowing the affective representation to focus on non-cognitive modulation.}
\label{fig:1}
\end{figure}

The main contributions of this paper are as follows.

\begin{enumerate}
\item We identify the affect contamination problem in affective cognitive diagnosis based on the observation that affective representations may absorb unmodeled cognitive residuals and become entangled with item, concept, and ability bias. This work complements existing affect-aware cognitive diagnosis studies, which focus on prediction improvement while paying less attention to representation contamination \cite{ref17}, \cite{ref21}.
\item We propose an ability-residual decoupling architecture that separates item, student, concept, student-concept personalized, and low-rank student-item interaction residuals from affective modeling. This strategy allows the affective module to focus more on non-cognitive factors.
\item We construct a Q-matrix-masked concept residual attention mechanism, which adaptively aggregates concept-shared residuals and student-concept personalized residuals as interpretable concept values within the item-related concept set. This approach improves the accuracy and interpretability of residual modeling.
\item We validate the proposed framework on ASSIST2017, ASSIST2012, ASSIST2009, and Junyi using multiple cognitive diagnosis backbones, including DINA, IRT, MIRT, NCDM, RCD, and SCD. Our method's effectiveness is systematically evaluated on response prediction, affect prediction, and representation decoupling through ablation experiments, residual group analysis, affect contamination diagnostics, and representation visualization \cite{ref1}, \cite{ref5}, \cite{ref9,ref10,ref11,ref12}, \cite{ref21}.
\end{enumerate}
\section{Related Work}

\subsection{Cognitive Diagnosis and Affect-Aware Extensions}

CD models estimate students' mastery of fine-grained knowledge concepts or cognitive attributes from response logs, supporting personalized instruction, formative assessment, and adaptive testing \cite{ref2,ref3,ref4}. Early psychometric and cognitive diagnosis models have highly interpretable structures. Classic CDMs include DINA, IRT, and MIRT. Specifically, DINA relies on the Q-matrix to represent discrete item-concept requirements and uses item parameters such as guessing and slipping to explain response noise beyond mastery states \cite{ref1}, \cite{ref8}. Besides, IRT models response probability as a function of continuous student ability and item difficulty \cite{ref9}, while MIRT extends unidimensional ability to a multidimensional latent vector for multi-skill or multi-concept assessment \cite{ref10}. These models preserve clear educational-measurement meanings among students, items, and concepts, although parametric assumptions can limit their expressive power.

As researchers have accumulated large-scale response data on online education platforms, deep learning has been incorporated into cognitive diagnosis, giving rise to neural CDMs. An NCDM combines response probabilities with neural networks while retaining interpretable variables and monotonicity constraints \cite{ref5}. In this case, RCD models structural relations among students, items, and concepts, whereas SCD introduces self-supervised signals to alleviate sparsity and label deficiency \cite{ref11}, \cite{ref12}. Higher-order and global-local models further use hierarchical structures, attention, or local concept relations \cite{ref7}, \cite{ref13}. These developments strengthen the representation of cognitive structure through more expressive student-, item-, and concept-level interactions, but their primary modeling target remains cognitive ability or mastery.

Observed responses are also affected by non-cognitive factors. Indeed, achievement emotions, confusion, frustration, and boredom can influence motivation, engagement, and performance \cite{ref14,ref15,ref16}. At the same time, affect-detection and affect-aware tutoring studies provide empirical and computational support for incorporating these factors into learner modeling \cite{ref17,ref18,ref19,ref20}. Affect-aware cognitive diagnosis extends conventional CDMs by estimating affective states and uses these estimations to modulate guessing and slipping \cite{ref21}. When external affect labels are unavailable, response-conditioned contrastive learning can provide weak supervision for learning response-derived affective representations \cite{ref22}, \cite{ref23}.

Although conventional and affect-aware CDMs improve cognitive-structure modeling and non-cognitive modulation, respectively, existing affect-aware extensions assume that the cognitive backbone has already explained stable student-, item-, and concept-related variation sufficiently \cite{ref5}, \cite{ref11}, \cite{ref12}, \cite{ref21}. In real educational data, item calibration bias, concept-level systematic bias, personalized student-concept deviations, and latent student-item matching may remain \cite{ref7}, \cite{ref24}, \cite{ref25}. Notably, if these residuals are not explicitly modeled, the affective branch may absorb them and act as a generic error compensator. In this case, response improvements alone cannot establish that the learned affective representation is aligned with non-cognitive states \cite{ref17}, \cite{ref21}. Motivated by this gap, this study assigns stable cognitive residual variation to a dedicated ability-residual pathway before affective modulation.

\subsection{Representation Decoupling}

Representation learning aims to capture factors that are useful for downstream tasks \cite{ref26}. Interestingly, a representation that improves prediction is not necessarily aligned with the target factor. Indeed, disentangled representation learning reveals that learned factors depend on task objectives, supervision signals, and structural inductive biases. Without appropriate inductive biases, general learning objectives cannot guarantee that a model will recover the expected factors \cite{ref27}. This limitation is especially important when multiple latent factors are statistically correlated, because a model may encode shortcut features or spurious correlations that are useful for prediction but do not correspond to the intended factor \cite{ref28}.

Representation decoupling overcomes this limitation by encouraging different latent components to carry different types of information. Depending on the task, decoupling can be introduced through architectural separation, factor-specific supervision, contrastive objectives, or adversarial constraints. Although these mechanisms do not guarantee complete independence, they reduce unnecessary information leakage and improve interpretability. Typical diagnostic tools are linear probes, which test whether a representation contains information about a factor that the model intends to separate \cite{ref29}. Although these do not enforce decoupling, they provide empirical evidence about whether the intended separation has been achieved. This connection motivates the leakage probes used in this work to evaluate cognitive information remaining in affective representations.

\subsection{Residual Modeling}

Residual learning is a general modeling concept widely used to let models learn the remaining function relative to a base mapping \cite{ref31}. Bias correction and low-rank interaction modeling are widely used in recommender systems and educational prediction tasks to explain stable individual differences and sparse interaction structures \cite{ref24}, \cite{ref25}, \cite{ref32,ref33,ref34}. Matrix factorization recommender models typically model user bias, item bias, and low-rank user-item interactions to explain personalized differences that the global mean and the main model cannot capture \cite{ref24}, \cite{ref25}, \cite{ref32}. This concept has a natural interpretation when transferred to educational data: student bias corresponds to a student's overall response tendency or baseline ability offset, item bias corresponds to item difficulty calibration error, and low-rank student-item interactions correspond to wording familiarity, problem-solving strategy matching, hidden skill requirements, or other unobserved matching relations. Similarly, response prediction in knowledge tracing and online assessment systems involves sparse interactions among students, items, and exercise records. Therefore, it requires additional structures to characterize individual differences and historical response patterns \cite{ref4}, \cite{ref6}, \cite{ref33}, \cite{ref34}.

From this decoupling perspective, residual modeling offers the structural mechanism for assigning stable cognitive variation to a dedicated pathway rather than to the affective representation. In affective cognitive diagnosis, ability residuals are not merely numerical corrections to response logits but allocate modeling roles. Thus, residual patterns that recur with students, items, concepts, or student-item combinations should be explained mainly by the ability-residual pathway. In contrast, the affective module effectively handles fluctuations associated with transient states, engagement, or affective changes. Accordingly, the ability-residual module includes student, item, concept-shared, student-concept personalized, and low-rank student-item components. Hence, residual modeling supports representation decoupling in addition to increasing model capacity.

\section{Problem Formulation}

Let $\mathcal{S}$ be the set of students, be $\mathcal{I}$ the set of items, and $\mathcal{K}$ the set of concepts. For a student-item interaction $(s,i)$, where $s\in\mathcal{S}$ and $i\in\mathcal{I}$, the concept association vector of item $i$ is denoted by $\mathbf{q}_i\in\{0,1\}^{K}$, and the response label is $y_{si}\in\{0,1\}$. If affect labels or affect vectors are available in the dataset, they are denoted by $\mathbf{a}_{si}$.

An observed response may reflect a latent cognitive state $\mathbf{c}_{si}$, a latent affective state $\mathbf{a}_{si}$, or their joint influence \cite{ref21}. Let $p_{si}$ denote the probability of a correct response:

\begin{equation}
p_{si}=P(y_{si}=1\mid \mathbf{c}_{si},\mathbf{a}_{si},i,\mathbf{q}_i)
\end{equation}

A conventional cognitive diagnosis model learns the following response probability:

\begin{equation}
\hat{y}_{si}=P(y_{si}=1\mid s,i,\mathbf{q}_i).
\end{equation}

Affective cognitive diagnosis further introduces an affective representation $\mathbf{a}_{si}$, which is usually written as

\begin{equation}
\hat{y}_{si}=f_{\theta}(s,i,\mathbf{q}_i,\mathbf{a}_{si}).
\end{equation}

However, when the base cognitive diagnosis backbone cannot sufficiently explain cognitive factors, the learned affective representation may contain the target non-cognitive state and the incorrectly absorbed cognitive residual, which is formulated as:

\begin{equation}
\mathbf{a}_{si}=\mathbf{a}^{affect}_{si}+\Delta^{cog}_{si},
\end{equation}

where $\mathbf{a}^{affect}_{si}$ is the non-cognitive component related to affective state and $\Delta^{cog}_{si}$ denotes cognitive residual variation that enters the affective space, without being explained by the base backbone, such as variation associated with item calibration bias, concept bias, or student-concept deviations.

Throughout this paper, \emph{cognitive residuals} refer to stable cognitive variation not explained by the base cognitive diagnosis backbone. In contrast, the \emph{ability residual} $r^{ability}_{si}$ is the structured, model-estimated correction introduced to capture such variation. \emph{Ability residual} is used in a broad diagnostic sense and is not limited to a student-specific ability offset, incorporating item, concept, student-concept, and low-rank student-item components.

Besides, $\Delta^{cog}_{si}$ is a conceptual latent variation rather than an observed training target. By contrast, $r^{ability}_{si}$ is a structured correction estimated by the model from the available data and learning objectives. The two quantities are not equivalent, and response labels alone cannot uniquely identify their decomposition or recover either quantity as ground truth.

This paper aims to explicitly model the ability residual $r^{ability}_{si}$ so that the final response probability satisfies:

\begin{equation}
\hat{y}_{si}=f_{\theta}(s,i,\mathbf{q}_i,\mathbf{a}^{affect}_{si},r^{ability}_{si}),
\end{equation}

and to reduce the entanglement between $\mathbf{a}^{affect}_{si}$ and $\Delta^{cog}_{si}$ through a structured allocation of modeling roles. In other words, the ability residual module explains unmodeled cognitive residual variation, while the affective module explains how non-cognitive states modulate guessing, slipping, and response tendency.

\section{Method}

\subsection{Overall Framework}

The proposed framework has three core components: a base cognitive diagnosis backbone, an ability residual module, and an affective module. The base backbone $F_{CD}$ outputs a base response probability according to student ability, item difficulty, item discrimination, and the Q-matrix:

\begin{equation}
y^{base}_{si}=F_{CD}(\mathbf{e}_s,\mathbf{d}_i,\mathbf{d}^{disc}_i,\mathbf{q}_i),
\end{equation}

where $\mathbf{e}_s$ denotes student ability or mastery state, $\mathbf{d}_i$ is item difficulty or concept difficulty, $\mathbf{d}^{disc}_i$ denotes item discrimination, and $\mathbf{q}_i$ denotes the concepts involved in the item. DINA, IRT, MIRT, NCDM, RCD, or SCD can replace this backbone.

Then, the ability residual module introduces $r^{ability}_{si}$ to capture cognitive residual variation not explained by the base backbone and applies this correction on the logit scale:

\begin{equation}
z^{ability}_{si}=\operatorname{logit}(y^{base}_{si})+r^{ability}_{si},
\qquad
y^{*}_{si}=\sigma(z^{ability}_{si}).
\end{equation}

Finally, the affective module generates an affective representation and modulates the residual-corrected response probability through the guess/slip mechanism. Unlike directly concatenating affective representations into the backbone, this structure explicitly distinguishes ability residuals from affective factors. Therefore, it is consistent with the modeling assumption on affect contamination presented in this paper.

\begin{figure}[htbp]
\centering
\includegraphics[width=0.96\linewidth]{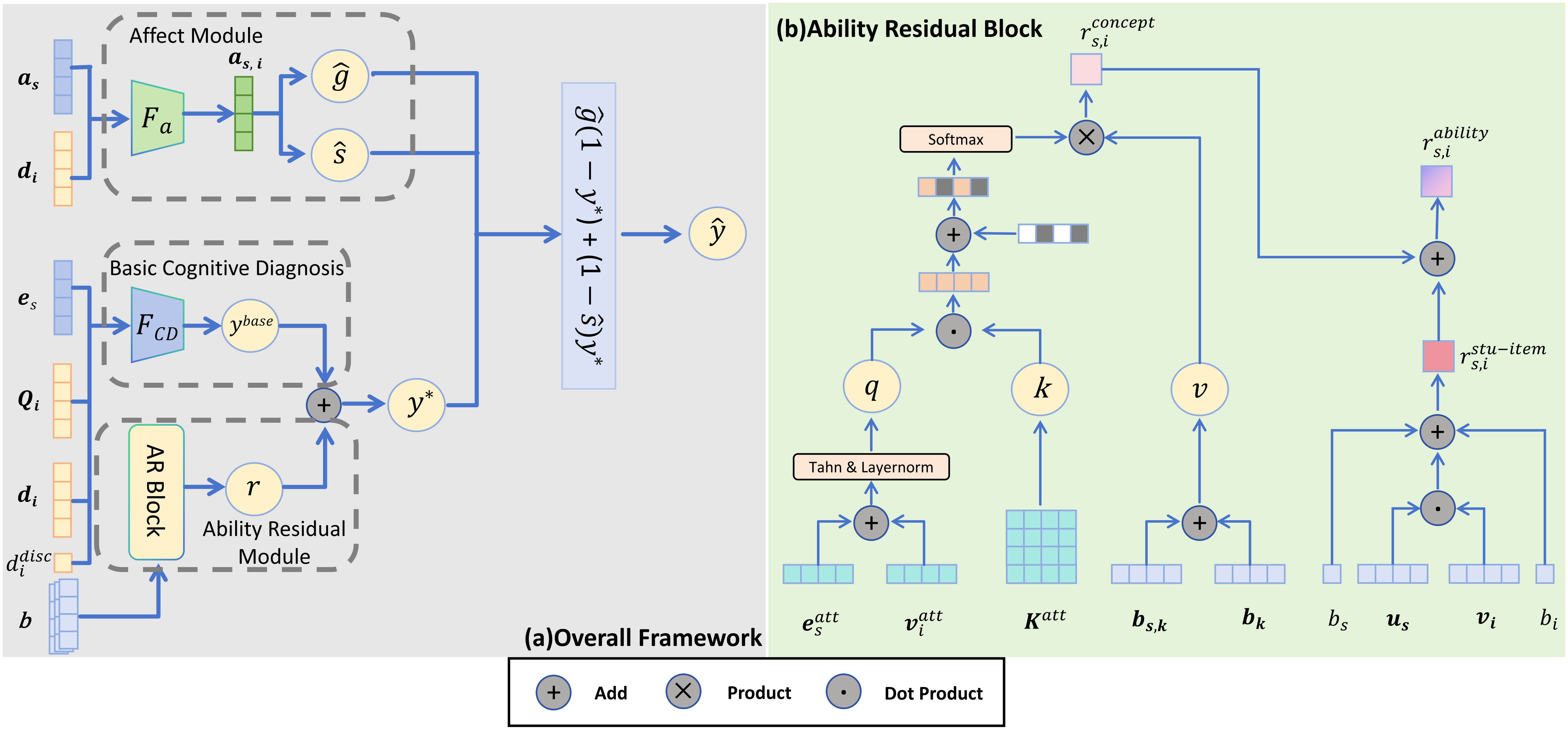}
\caption{Overall ARACD algorithmic framework and ability residual module.}
\label{fig:2}
\end{figure}

\subsection{Base Cognitive Diagnosis Backbone}

The base cognitive diagnosis backbone is treated as a replaceable module. Considering NCDM as an example, the student knowledge mastery, item concept difficulty, and item discrimination vectors jointly form an interaction representation:

\begin{equation}
\mathbf{x}_{si}=\mathbf{d}^{disc}_i\odot(\mathbf{e}_s-\mathbf{d}^{kc}_i)\odot\mathbf{q}_i,
\end{equation}

\begin{equation}
y^{base}_{si}=\sigma(MLP(\mathbf{x}_{si})).
\end{equation}

where $\odot$ denotes element-wise multiplication, $\mathbf{d}^{disc}_i$ is item discrimination, and $\mathbf{d}^{kc}_i$ is the item's concept difficulty representation. For IRT and MIRT, the backbone formulation reduces to a matching function between unidimensional or multidimensional latent ability and item parameters. Accordingly, for DINA, the backbone estimates response probability based on whether the student has mastered all concepts required by the item and on guess/slip parameters. The proposed framework is designed as a plug-in that can be attached to different cognitive diagnosis models.

\subsection{Ability Residual Modeling}

The ability residual explains cognitive residual variation not captured by the base backbone. However, to improve modeling stability and interpretability, the ability residual is divided into a student-item residual layer and a concept residual layer. The former layer models overall offsets and personalized interactions that do not depend on the concept structure, including student global bias, item global bias, and low-rank student-item latent interactions. The latter layer uses Q-matrix constraints to aggregate concept-shared residuals and student-concept personalized residuals within the concepts related to the item. This division separates non-concept-level response bias from concept-level knowledge bias, making the residual structure more compact and interpretable.

This modeling strategy absorbs cognitive residuals from non-affective sources because each residual term corresponds directly to a common stable error source in educational measurement. Specifically, item global bias absorbs item difficulty calibration errors; student global bias absorbs a student's overall response tendency or baseline ability offset; concept-shared residuals absorb systematic prediction bias on certain concepts; student-concept personalized residuals absorb deviations in a student's mastery of specific concepts; and low-rank student-item interactions absorb student-item-specific errors caused by wording familiarity, strategy matching, or hidden skill requirements. These factors commonly appear with the structure of students, items, or concepts, rather than being determined instantaneously by the affective state in a single interaction. Therefore, explicitly placing them in the ability residual pathway is a more appropriate modeling pathway for cognitive bias not explained by the base backbone.

During training, the ability residual corrects the base backbone output on the logit scale and is connected to the final response loss. If a certain type of error repeatedly appears for the same student, item, concept, or similar student-item combinations, gradients update the corresponding residual parameters and assign this stable error to the ability residual pathway. In contrast, the affective module modulates the final response probability through guess/slip and is more suitable for explaining response fluctuations induced by non-cognitive states. This structural division does not assume that the two are completely separable. Instead, it reduces the pressure on the affective branch to absorb cognitive bias through parameter allocation and educational structural constraints, thereby mitigating affect representation contamination.

\begin{equation}
r^{ability}_{si}=w_{\mathrm{SI}}r^{stu\text{-}item}_{si}+w_{\mathrm{C}}r^{concept}_{si},
\end{equation}

where $r^{stu\text{-}item}_{si}$ is the student-item residual layer, $r^{concept}_{si}$ is the concept residual layer, and $w_{\mathrm{SI}}$ and $w_{\mathrm{C}}$ are the scaling coefficients of the two layers. Here $w_{\mathrm{SI}}$ is an overall weight for the student-item residual layer rather than a parameter that varies separately with sample $(s,i)$. In implementation, it can be set as either a learnable parameter or fixed to 1.

All $b$ terms in the ability residual are internal learnable parameters rather than externally provided labels or manually calculated rules. During training, the model retrieves these residual terms from the corresponding parameter tables according to student, item, and concept IDs, and updates them end-to-end through the response prediction loss and affect representation learning loss. To avoid ambiguity in notation, the main symbols used in ability residual decomposition and concept attention are summarized below.

\begin{table*}[htbp]
\centering
\caption{Main notation used in the ability residual decomposition and concept residual attention.}\label{tab:1}
\small
\renewcommand{\arraystretch}{1.15}
\begin{tabularx}{\textwidth}{@{}lX@{}}
\toprule
Symbol & Meaning \\
\midrule
$s$ & The student in the considered student--item pair \\
$i$ & The item in the considered student--item pair \\
$k$ & The $k$-th concept \\
$\mathbf{q}_i$ & Concept mask of item $i$ \\
$\mathcal{K}_i$ & The set of concepts involved in item $i$, $\mathcal{K}_i=\{k:q_{ik}=1\}$ \\
$b_s$ & Stable offset indicating that student $s$ is more or less likely than the overall level to answer correctly \\
$b_i$ & Difficulty calibration error or item-specific systematic offset of item $i$ \\
$b_k,\mathbf{b}$ & Systematic residual shared by concept $k$ across all students and items, and the vector of all concept-shared residuals \\
$b_{s,k},\mathbf{b}^{concept}_s$ & Personalized mastery deviation of student $s$ on concept $k$ relative to his or her overall ability, and the vector of all concept-specific personalized residuals for student $s$ \\
$v_{s,k},\mathbf{v}_s$ & Composite concept residual value for student $s$ on concept $k$, $v_{s,k}=b_k+b_{s,k}$ \\
$\mathbf{u}_s,\mathbf{v}_i$ & Latent student-item matching relation, whose inner product approximates the complete interaction residual matrix \\
$\gamma_{SI}$ & Strength coefficient of the low-rank student-item interaction residual \\
$\mathbf{e}^{att}_s$ & Student-specific embedding used to generate the student--item-pair-conditioned query \\
$\mathbf{v}^{att}_i$ & Item-specific embedding used to generate the student--item-pair-conditioned query \\
$\mathbf{q}^{att}_{si}$ & Student--item-pair-conditioned query, $\mathbf{q}^{att}_{si}=LN(\tanh(\mathbf{e}^{att}_s+\mathbf{v}^{att}_i))$ \\
$\mathbf{k}^{att}_k,\mathbf{K}^{att}$ & Attention key of the $k$-th concept and all concept keys \\
$e_{sik},\mathbf{e}_{si}$ & Concept attention scores before Q-mask application \\
$m_{ik},\mathbf{m}_i$ & Q-matrix mask that restricts attention to item-related concepts \\
$\alpha_{sik},\boldsymbol{\alpha}_{si}$ & Q-masked concept attention weights, $\alpha_{sik}=\operatorname{softmax}_{k}(e_{sik}+m_{ik})$ \\
$r^{att}_{si}$ & Output of the concept residual layer, $r^{att}_{si}=\sum_k\alpha_{sik}v_{s,k}$ \\
\bottomrule
\end{tabularx}
\end{table*}

These residual biases are usually initialized to 0 so that the model initially degenerates as much as possible to the base cognitive diagnosis backbone. Low-rank interaction vectors and attention vectors are initialized with small random values. As training proceeds, if a certain type of student, item, or concept exhibits stable errors under the base backbone, gradients push the corresponding residual parameters away from 0, absorbing this part of cognitive bias into the ability residual pathway.

\subsubsection{Student-Item Residual Layer}

The student-item residual layer is formulated as:

\begin{equation}
r^{stu\text{-}item}_{si}=r^{student}_{s}+r^{item}_{i}+r^{interaction}_{si},
\end{equation}

which absorbs overall offsets from students and items and models fine-grained latent student-item matching. Specifically,

\begin{equation}
r^{student}_{s}=b_s,
\qquad
r^{item}_{i}=b_i,
\qquad
r^{interaction}_{si}=\gamma_{SI}\frac{\langle \mathbf{u}_s,\mathbf{v}_i\rangle}{\sqrt{d}}.
\end{equation}

where $b_s$ and $b_i$ correspond to the student global bias and item global bias in the table above, $\mathbf{u}_s$ and $\mathbf{v}_i$ are the low-rank interaction vectors of the student and item, respectively, $d$ is the low-rank dimension, and $\gamma_{SI}$ is the strength coefficient of the low-rank student-item interaction residual, corresponding to student\_item\_rank\_scale. When $\gamma_{SI}$ is small, the low-rank interaction term has a weak influence on the final residual. When $\gamma_{SI}$ is large, the model makes stronger use of personalized student-item matching to correct predictions from the base backbone. The main experiments use $\gamma_{SI}=1.5$ by default. To avoid the high parameter cost and overfitting risk of directly modeling the complete student-item interaction matrix, we use the low-rank decomposition $\mathbf{u}_s^\top \mathbf{v}_i$ to represent personalized residuals at the student-item level. This design reduces computational and storage overhead while imposing a low-rank constraint on the interaction structure, thereby improving generalization.

\subsubsection{Concept Residual Layer}

The concept residual layer is defined as:

\begin{equation}
r^{concept}_{si}=r^{att}_{si},
\end{equation}

and models systematic concept-level bias and student-concept personalized bias. This layer introduces the Q-matrix-constrained concept residual attention mechanism to aggregate item-related concepts through attention weighting, as detailed in Section 4.4. Specifically,

\begin{equation}
r^{att}_{si}=\sum_{k=1}^{K}\alpha_{sik}(b_k+b_{s,k}),
\end{equation}

where $b_k$ is the shared concept residual bias of the $k$-th concept, representing systematic bias of that concept across all students and items; $b_{s,k}$ is the personalized residual bias of student $s$ on concept $k$, representing the student's mastery deviation on that specific concept. Their sum provides the concept-layer value, which denotes the composite concept residual of the current student on that concept. $\mathcal{K}_i=\{k:q_{ik}=1\}$ is the set of concepts involved in item $i$, and the Q-matrix mask zeroes out the attention weights of irrelevant concepts.

This decomposition has clear interpretability. Indeed, in the student-item residual layer, $b_s$ and $b_i$ absorb overall student and item offsets, respectively, and the low-rank interaction term $r^{interaction}_{si}$reflect latent matching relations induced by item wording, problem-solving strategy, or hidden skill requirements. In the concept residual layer, $b_k+b_{s,k}$ explains systematic concept bias and student-concept personalized mastery deviations. This two-layer architecture facilitates ablation analysis and allows direct statistics on the contributions of different residual sources to the final response probability.

\subsection{Q-Matrix-Constrained Concept Residual Attention}

Next, we introduce a Q-matrix-constrained concept residual attention mechanism into the concept residual layer $r^{concept}_{si}$. Among the concepts associated with item $i$, the mechanism adaptively determines the relative contribution of each explicit concept residual to the concept-level correction for student $s$'s predicted response to item $i$. Specifically, a student--item-pair-conditioned query is constructed by combining a student-specific and an item-specific attention embedding. Its compatibility with each candidate concept key provides the basis for residual allocation: a higher compatibility score assigns a larger relative weight to the corresponding concept residual. The Q-matrix further restricts this allocation to concepts required by the item. It should be noted that attention is used only for concept-layer aggregation, not for the student-item residual layer.

The proposed module is a structured residual aggregation rather than a conventional hidden-state attention \cite{ref30}. It does not use attention to create a new hidden value. Still, it leaves each scalar $v_{s,k}$ untransformed and uses attention only to determine its relative allocation. The resulting correction remains directly decomposable into shared and student-specific concept residuals.

During implementation, we directly learn concept key vectors, whose role is equivalent to $\mathbf{k}^{att}_k=\mathbf{W}_k\mathbf{c}_k$ in the theoretical expression. Specifically, the model does not explicitly store $\mathbf{W}_k$ and $\mathbf{c}_k$ as two separate objects but combines their product into a learnable vector $\mathbf{k}^{att}_k$. We use the symbols in the table above and describe the computation for a student--item pair $(s,i)$.

First, the query for the student--item pair is constructed from the student and the item:

\begin{equation}
\mathbf{q}^{att}_{si}=LN\left(\tanh(\mathbf{e}^{att}_s+\mathbf{v}^{att}_i)\right),
\qquad
\mathbf{q}^{att}_{si}\in\mathbb{R}^{d_{att}}.
\end{equation}

where $\mathbf{e}^{att}_s$ and $\mathbf{v}^{att}_i$ are obtained from student and item embedding tables, respectively, and both are vectors of length $d_{att}$. $\tanh(\cdot)$ denotes the hyperbolic tangent activation function, which compresses the combined student-side and item-side query information into a bounded nonlinear space. Moreover, $LN(\cdot)$ denotes Layer Normalization, which stabilizes the scale of the query vector and avoids unstable fluctuations in subsequent dot-product scores caused by excessively large vector norms. For each concept $k$, the model directly learns a key vector:

\begin{equation}
\mathbf{k}^{att}_{k}\in\mathbb{R}^{d_{att}},
\qquad
\mathbf{K}^{att}=[\mathbf{k}^{att}_{1};\cdots;\mathbf{k}^{att}_{K}]\in\mathbb{R}^{K\times d_{att}}.
\end{equation}

Therefore, for a student--item pair $(s,i)$, the unmasked attention score on concept $k$ is defined as:

\begin{equation}
e_{sik}=\frac{(\mathbf{q}^{att}_{si})^T\mathbf{k}^{att}_{k}}{\sqrt{d_{att}}},
\qquad
e_{sik}\in\mathbb{R}.
\end{equation}

By concatenating the scores of all concepts, vector $\mathbf{e}_{si}\in\mathbb{R}^{K}$ is obtained. To ensure that attention is allocated only over item-related concepts, the item-level Q-matrix mask $\mathbf{m}_i\in\mathbb{R}^{K}$ is introduced:

\begin{equation}
m_{ik}=\begin{cases}
0, & q_{ik}=1,\\
-\infty, & q_{ik}=0,
\end{cases}
\end{equation}

where $q_{ik}$ is the 0/1 mark of item $i$ on concept $k$ in the Q-matrix. The final attention weight is:

\begin{equation}
\alpha_{sik}=\operatorname{softmax}_{k}(e_{sik}+m_{ik}),
\qquad
\boldsymbol{\alpha}_{si}\in\mathbb{R}^{K}.
\end{equation}

The concept residual value remains an interpretable scalar. For concept $k$, $v_{s,k}$ is a scalar and for all concepts it forms the vector $\mathbf{v}_s=[v_{s,1},\ldots,v_{s,K}]\in\mathbb{R}^{K}$. In the proposed solution, the shared concept residual vector $\mathbf{b}\in\mathbb{R}^{K}$ is added to the student-concept residual vector $\mathbf{b}^{concept}_s\in\mathbb{R}^{K}$:

\begin{equation}
v_{s,k}=b_k+b_{s,k},
\qquad
\mathbf{v}_s\in\mathbb{R}^{K}.
\end{equation}

The Q-matrix defines the candidate set $\mathcal{K}_i$, whereas the query-key compatibility scores determine the relative allocation within that set. Thus, $\alpha_{sik}$ reflects student--item-pair--concept compatibility rather than directly the magnitude of $v_{s,k}$, which is used only after the weights are obtained. Therefore, it is interpreted as a residual-allocation coefficient, and not as a mastery probability or causal importance score.

Concept residual aggregation is a weighted sum along the concept dimension, $K$:

\begin{equation}
r^{att}_{si}=\sum_{k=1}^{K}\alpha_{sik}v_{s,k},
\qquad
r^{att}_{si}\in\mathbb{R}.
\end{equation}

For a single-concept item, the masked softmax gives $\alpha_{sik^*}=1$ and $r^{att}_{si}=v_{s,k^*}$, and adaptive allocation is activated only for multi-concept items.

During implementation, the shared concept term and the student-concept personalized term can be separately summarized:

\begin{equation}
r^{shared}_{si}=\sum_{k=1}^{K}\alpha_{sik}b_k,
\qquad
r^{student\text{-}concept}_{si}=\sum_{k=1}^{K}\alpha_{sik}b_{s,k}.
\end{equation}

Their sum gives the concept residual layer $r^{concept}_{si}=r^{att}_{si}$, which is then combined with the student-item residual layer $r^{stu\text{-}item}_{si}$ to form the final ability residual.

\subsection{Affective Module}

The affective module learns the non-cognitive state of a student in the current interaction. Given the student affective representation $\mathbf{a}_s$ and the item concept difficulty representation $\mathbf{d}^{kc}_i$, the interaction-level affective representation is formulated as:

\begin{equation}
\mathbf{a}_{si}=g_{\theta}(\mathbf{a}_s,\mathbf{d}^{kc}_i),
\end{equation}

where $\mathbf{a}_s$ is a learnable student-level affective representation obtained by student ID lookup, $\mathbf{d}^{kc}_i$ is the item concept difficulty representation in the base cognitive diagnosis backbone and provides the knowledge requirement context of the current item, and $g_{\theta}(\cdot)$ is a learnable mapping function used to generate the affective representation of the current student-item interaction. These parameters are learned end-to-end through the training objective, rather than manually specified by external affective rules.

Let guess and slip be directly determined by the affective representation:

\begin{equation}
g_{si}=\sigma(\mathbf{w}_g^T\mathbf{a}_{si}+b_g),
\end{equation}

\begin{equation}
s_{si}=\sigma(\mathbf{w}_{slip}^T\mathbf{a}_{si}+b_{slip}).
\end{equation}

where $\mathbf{w}_g$ and $b_g$ are the linear weight and bias of the guess branch, and $\mathbf{w}_{slip}$ and $b_{slip}$ are the linear weight and bias of the slip branch. The terms $b_g$ and $b_{slip}$ belong only to the intercepts of the affective modulation branch. They are not part of the student, item, or concept residuals in the ability residual module.

The final response probability is presented in Eq. (26) and indicates that when $y^{*}_{si}$ is low, the guess term may increase the probability of a correct response; when $y^{*}_{si}$ is high, the slip term may decrease the probability of a correct response. The final probability is therefore jointly determined by the residual-corrected prediction, the guess term, and the slip term.

\begin{equation}
\hat{y}_{si}=(1-s_{si})y^{*}_{si}+g_{si}(1-y^{*}_{si}).
\end{equation}

\subsection{Affective Representation Learning With and Without Affect Labels}

Depending on whether real affect labels are available in the dataset, ACD follows two affect-aware training strategies. When the training data contain external affect annotations, supervised affect prediction is used. When real affect annotations are unavailable, contrastive affect representation learning is used by constructing positive and negative samples based on response labels. Let $\mathcal{B}$ be the current training batch, $\mathbf{a}^{gt}_i\in[0,1]^{d_{aff}}$ be the real affect vector of sample $i$, and $\hat{\mathbf{a}}_i$ be the model's affective representation. Given that the original ACD treats affective states as continuous values, the supervised affect perception module uses mean squared error:

\begin{equation}
\mathcal{L}_{a}
=
\frac{1}{|\mathcal{B}|}
\sum_{i\in\mathcal{B}}
\left\|\hat{\mathbf{a}}_i-\mathbf{a}^{gt}_i\right\|_2^2.
\end{equation}

This loss corresponds to the affective perception loss in ACD and makes the affective representation induced by student affective traits and item difficulty close to the external affect annotations. In ARACD, $\hat{\mathbf{a}}_i$ still participates in final response prediction through guess/slip modulation. Besides, the ability residual has no independent supervision label, as its parameters are updated jointly by backpropagation through the response prediction path and the affect prediction path.

For CACD/ARCACD settings without real affect labels, this study follows the contrastive affect perception idea in the original ACD and replaces $\mathcal{L}_{a}$ with a contrastive loss. Its basic assumption is that student-item interactions with the same response outcome are more likely to correspond to similar affective states. In contrast, interactions with different response outcomes are more likely to correspond to different affective states. For any sample $i$ in a batch, the positive and negative sample sets are constructed as:

\begin{equation}
\mathcal{P}_i=\{j\in\mathcal{B}:j\ne i,\ y_j=y_i\},
\qquad
\mathcal{N}_i=\{k\in\mathcal{B}:y_k\ne y_i\}.
\end{equation}

where $y_i\in\{0,1\}$ is the response label of sample $i$. Let $\operatorname{sim}(\cdot,\cdot)$ denote cosine similarity between two affective vectors and $\tau$ be the temperature coefficient. Following CACD, the contrastive affective loss is defined as:

\begin{equation}
\mathcal{L}_{ca}
=
-\frac{1}{|\mathcal{I}|}
\sum_{i\in\mathcal{I}}
\frac{1}{|\mathcal{P}_i|}
\sum_{j\in\mathcal{P}_i}
\log
\frac{
\exp\left(\operatorname{sim}(\hat{\mathbf{a}}_i,\hat{\mathbf{a}}_j)/\tau\right)
}{
\sum_{k\in\mathcal{N}_i}\exp\left(\operatorname{sim}(\hat{\mathbf{a}}_i,\hat{\mathbf{a}}_k)/\tau\right)
},
\end{equation}

where

\begin{equation}
\mathcal{I}=\{i\in\mathcal{B}:|\mathcal{P}_i|>0,\ |\mathcal{N}_i|>0\}
\end{equation}

is the set of valid anchors. This training strategy does not introduce real affect labels for individual samples. Instead, it uses weak positive and negative relations formed by response labels to constrain the affective representation space, thereby providing latent affective cues for guess/slip modulation.

\subsection{Training Objective}

The response prediction loss follows the binary cross-entropy used in ACD. For a batch $\mathcal{B}$, let $\hat{y}_i$ be the final predicted probability and $y_i$ the true response label. The loss is formulated as:

\begin{equation}
\mathcal{L}_{CDM}
=
-\frac{1}{|\mathcal{B}|}
\sum_{i\in\mathcal{B}}
\left[y_i\log \hat{y}_i+(1-y_i)\log(1-\hat{y}_i)\right].
\end{equation}

In supervised ACD/ARACD settings with real affect labels, response prediction loss and affect prediction loss are optimized jointly:

\begin{equation}
\mathcal{L}_{sup}=\mathcal{L}_{CDM}+\lambda_a\mathcal{L}_{a},
\end{equation}

where $\lambda_a$ is the trade-off coefficient of the affect prediction loss. $\mathcal{L}_{CDM}$ constrains the final response probability, and $\mathcal{L}_{a}$ constrains the predicted affective vector to align with the external affect annotations.

In CACD/ARCACD settings without real affect labels, $\mathcal{L}_{ca}$ replaces the supervised affect MSE, and the training objective becomes:

\begin{equation}
\mathcal{L}_{unsup}=\mathcal{L}_{CDM}+\lambda_c\mathcal{L}_{ca},
\end{equation}

where $\lambda_c$ is the trade-off coefficient of the contrastive affect perception loss. This setting does not compute $\mathcal{L}_{a}$. The affective representation influences response prediction through the guess/slip branch and receives weak supervision from response labels through $\mathcal{L}_{ca}$. Apart from the supervised affect loss or contrastive affect loss, this study does not introduce additional decorrelation losses or residual losses. Necessary control of parameter scale is handled by optimizer weight decay or parameter clipping, but is not reported as an independent training objective.

\section{Experimental Setup}

\subsection{Research Questions}

The study is organized around the following research questions.

\textbf{RQ1: Can ability-residual decoupling stably improve response prediction performance?} This work compares Base, ACD/CACD, and ARACD/ARCACD on ASSIST2017, ASSIST2012, ASSIST2009, and Junyi, and examines whether ACC, AUC, RMSE, and MAE have consistent improvements across datasets and backbones.

\textbf{RQ2: Does the ability residual help affect alignment?} ACD and ARACD are compared using Affect RMSE and Affect MAE to test whether explicitly modeling cognitive residuals reduces the explanatory pressure on the affective branch caused by item, concept, and ability bias.

\textbf{RQ3: Which residual components account for the benefit of ability residuals?} Hierarchical ablations are conducted among Full, Only stu-item residual, Only concept residual, and w/o ability residual to evaluate the contributions of the student-item residual layer, the concept residual layer, and their combination to response prediction and affect prediction.

\textbf{RQ4: How do the attention dimension and low-rank dimension affect response prediction and affect alignment?} The concept attention dimension $d_{att}$ and the student-item low-rank dimension $rank$ are modified to examine the sensitivity of response metrics and affect metrics.

\textbf{RQ5: Do the two residual layers capture response-related structure rather than random perturbations?} The student-item residual and concept residual are visualized as a two-dimensional residual space to examine whether they form response-related patterns.

\textbf{RQ6: Can ability residuals mitigate affect contamination and improve model reliability?} Synthetic contamination probes, affect representation leakage diagnostics, and PCA visualization are used to analyze whether affective representations mix in cognitive bias information, and whether ability residuals can transfer this bias to the ability-residual pathway.

\textbf{RQ7: Can the proposed framework improve long-tail student diagnosis?} We further reanalyze low-activity students to examine whether ability residuals improve diagnosis for students with few interactions.

\subsection{Datasets}

The experiments use the educational datasets: ASSIST2017, ASSIST2012, ASSIST2009, and Junyi. All contain student response records, item IDs, concept annotations, and response correctness labels. ASSIST2017 and ASSIST2012 contain real affect labels and are used for supervised ACD/ARACD experiments. ASSIST2009 and Junyi do not use real affect supervision and are used for CACD/ARCACD experiments without affect labels. For each dataset, a fixed train/validation/test split ratio of 7:2:1 is used, while the data processing pipeline and evaluation protocol are consistent across all models.

\begin{center}
\begin{minipage}{\linewidth}
\centering
\refstepcounter{table}\label{tab:2}
{\small\textbf{Table \thetable}}\par
{\small Statistics of the four educational datasets.}\par\vspace{2pt}
\scriptsize
\setlength{\tabcolsep}{3pt}
\renewcommand{\arraystretch}{0.94}
\begin{tabular}{@{}cccc@{}}
\toprule
Dataset & Students & Items & Concepts \\
\midrule
ASSIST2017 & 1709 & 3162 & 102 \\
ASSIST2012 & 28998 & 47124 & 265 \\
ASSIST2009 & 4163 & 17746 & 123 \\
Junyi & 70627 & 2737 & 722 \\
\bottomrule
\end{tabular}
\end{minipage}
\end{center}

\subsection{Baseline Models and Compared Methods}

Six representative models are used as base backbones: DINA, IRT, MIRT, NCDM, RCD, and SCD to examine whether the ability residual framework can be attached as a general plug-in to different cognitive diagnosis backbones. These six backbones cover discrete cognitive diagnosis, unidimensional and multidimensional IRT, neural cognitive diagnosis, relational structure modeling, and self-supervised cognitive diagnosis. Therefore, these models allow us to evaluate the stability of the proposed method under different modeling assumptions and expressive capacities. In addition to base cognitive diagnosis backbones, ACD and CACD are used as affect-aware comparison methods to distinguish between "introducing an affective module" and "further introducing ability residuals".

\begin{center}
\begin{minipage}{\linewidth}
\centering
\refstepcounter{table}\label{tab:3}
{\small\textbf{Table \thetable}}\par
{\small Cognitive diagnosis backbones used in the experiments.}\par\vspace{2pt}
\scriptsize
\setlength{\tabcolsep}{2pt}
\renewcommand{\arraystretch}{1.02}
\begin{tabularx}{\linewidth}{@{}lX@{}}
\toprule
Backbone & Description \\
\midrule
DINA & A discrete cognitive diagnosis model that assumes students need to master all concepts required by an item to answer it correctly with high probability, and models guessing and slipping through guess/slip parameters. \\
IRT & A unidimensional item response theory model that estimates response probability based on the matching relation between student ability and item parameters, serving as a classical psychometric baseline. \\
MIRT & A multidimensional item response theory model that extends student ability into a multidimensional latent vector to characterize ability differences in multi-skill or multi-concept assessments. \\
NCDM & A neural cognitive diagnosis model that uses neural networks to fit response probabilities based on interpretable variables such as student mastery, item concept difficulty, and item discrimination. \\
RCD & A relation-aware cognitive diagnosis model that enhances interaction representations through relation structures among students, items, and concepts, and is used to evaluate the generalization of ability residuals on structured backbones. \\
SCD & A self-supervised cognitive diagnosis model that introduces self-supervised signals into diagnostic representations to improve robustness under sparse response scenarios. \\
\bottomrule
\end{tabularx}
\end{minipage}
\end{center}

\begin{center}
\begin{minipage}{\linewidth}
\centering
\refstepcounter{table}\label{tab:4}
{\small\textbf{Table \thetable}}\par
{\small Affect-aware comparison methods.}\par\vspace{2pt}
\scriptsize
\setlength{\tabcolsep}{2pt}
\renewcommand{\arraystretch}{1.04}
\begin{tabularx}{\linewidth}{@{}lX@{}}
\toprule
Method & Description \\
\midrule
ACD & A supervised affective cognitive diagnosis model that adds an affect-aware module to the base cognitive diagnosis backbone, uses real affect labels to constrain affective representations, and modulates response probability through guess/slip. \\
CACD & A contrastive affective cognitive diagnosis model for settings without real affect labels; it uses response labels to construct positive and negative sample relations for learning response-derived affective representations and modulates response prediction through these representations. \\
\bottomrule
\end{tabularx}
\end{minipage}
\end{center}

On ASSIST2017 and ASSIST2012, where real affect labels are available, we compare three methods for each backbone: Base uses only the base cognitive diagnosis backbone without an affective module or ability residual; ACD adds an affect-aware module to the same backbone and performs supervised affect prediction with real affect labels, but does not explicitly model ability residuals; ARACD adds the ability residual module and Q-matrix-constrained concept residual attention on top of ACD. This setting tests whether ability residuals can further improve response prediction and affect prediction beyond affect supervision.

On ASSIST2009 and Junyi, where real affect labels are not used, we compare Base, CACD, and ARCACD. CACD does not use per-sample real affect labels but constructs positive and negative sample relations from response labels for contrastive affective representation learning. ARCACD further introduces the ability residual module into CACD's contrastive affective representation learning framework. This setting tests whether ability residuals can still combine with response-derived affective representations and improve response prediction when external affect annotations are absent.

\subsection{Evaluation Metrics and Settings}

The main experiments use two types of metrics: response prediction metrics, including ACC, AUC, RMSE, and MAE, and affect alignment metrics, used only in supervised ACD/ARACD settings with real affect labels, including Affect RMSE and Affect MAE. For the former type, ACC denotes prediction accuracy after thresholding at 0.5; AUC measures whether the model assigns higher probabilities to correct responses; and RMSE and MAE measure probability error against the binary response label, with lower values being better.

All metrics measure the error between the predicted affective vectors and real affect labels, and all methods use the same data preprocessing, train/validation/test split, and evaluation scripts to ensure comparability. ASSIST2017 and ASSIST2012 use supervised affective learning, optimizing response prediction loss and real affect prediction loss. Additionally, ASSIST2009 and Junyi do not use real affect labels and adopt response-label-based contrastive affective representation learning. Except for hyperparameter sensitivity experiments, the main experiments keep the same training configuration within each dataset and backbone, select model checkpoints according to validation performance, and report final metrics on the test set. The experimental setup comprises a single NVIDIA GeForce RTX 3070 Ti GPU with 8 GB of memory. For model structural parameters, the main experiments use a concept attention dimension $d_{att}=16$, a student-item low-rank interaction dimension $rank=1024$, and a low-rank student-item residual strength coefficient $\gamma_{SI}=1.5$ by default. All models are optimized with Adam using a learning rate of $5\times10^{-4}$. The supervised experiments on ASSIST2017 and ASSIST2012 use a batch size of 512, train for at most 35 epochs with early-stopping patience values of 5 and 2, respectively, and optimize $\mathcal{L}_{CDM}+\mathcal{L}_{a}$. The contrastive experiments on ASSIST2009 and Junyi optimize $\mathcal{L}_{CDM}+\mathcal{L}_{ca}$, use a batch size of 256 and patience 5, and train for at most 20 and 30 epochs, respectively; RCD and SCD on Junyi use a batch size of 64 because of memory demand.

\section{Experimental Results and Analysis}

This section reports response prediction, affect alignment, ablation, hyperparameter sensitivity, residual maps, contamination diagnostics, long-tail results, and a student-level case study according to the research questions. Sections 6.1--6.7 correspond to RQ1--RQ7, respectively, while Section 6.8 provides supplementary student-level evidence.

\subsection{Response Prediction Results (RQ1)}

This section answers RQ1, namely whether ability-residual decoupling can stably improve response prediction performance under multi-dataset and multi-backbone settings. Table 5 reports the supervised affective learning results on ASSIST2017 and ASSIST2012. For each backbone, Base, ACD, and ARACD are retained, and four response metrics are reported: ACC, AUC, RMSE, and MAE. Table 6 lists the results without real affect labels on ASSIST2009 and Junyi, comparing Base, CACD, and ARCACD to test whether ability residuals can combine with response-label contrastive affective representation learning and bring stable gains.

\begin{table*}[htbp]
\centering
\caption{Response prediction results on ASSIST2017 and ASSIST2012.}\label{tab:5}
\scriptsize
\setlength{\tabcolsep}{2pt}
\renewcommand{\arraystretch}{1.12}
\resizebox{0.95\textwidth}{!}{%
\begin{tabular}{@{}cccccccccc@{}}
\toprule
\multirow{2}{*}{Backbone} & \multirow{2}{*}{Method} & \multicolumn{4}{c}{ASSIST2017} & \multicolumn{4}{c}{ASSIST2012} \\
\cmidrule(lr){3-6} \cmidrule(lr){7-10}
 &  & ACC $\uparrow$ & AUC $\uparrow$ & RMSE $\downarrow$ & MAE $\downarrow$ & ACC $\uparrow$ & AUC $\uparrow$ & RMSE $\downarrow$ & MAE $\downarrow$ \\
\midrule
\multirow{3}{*}{DINA} & Base & 0.6471 & 0.6950 & 0.4681 & 0.4383 & 0.7118 & 0.6923 & 0.4401 & 0.3654 \\
 & ACD & 0.7103 & 0.7760 & 0.4374 & 0.4008 & 0.7352 & 0.7397 & 0.4234 & 0.3489 \\
 & \textbf{ARACD} & \textbf{0.7390} & \textbf{0.8079} & \textbf{0.4202} & \textbf{0.3449} & \textbf{0.7494} & \textbf{0.7609} & \textbf{0.4139} & \textbf{0.3340} \\
\midrule
\multirow{3}{*}{IRT} & Base & 0.6582 & 0.7224 & 0.4668 & 0.4086 & 0.7286 & 0.7242 & 0.4289 & 0.3440 \\
 & ACD & 0.7152 & 0.7843 & 0.4338 & 0.3732 & 0.7401 & 0.7518 & 0.4194 & 0.3354 \\
 & \textbf{ARACD} & \textbf{0.7394} & \textbf{0.8082} & \textbf{0.4201} & \textbf{0.3442} & \textbf{0.7489} & \textbf{0.7602} & \textbf{0.4142} & \textbf{0.3334} \\
\midrule
\multirow{3}{*}{MIRT} & Base & 0.6801 & 0.7399 & 0.4661 & 0.4321 & 0.7355 & 0.7232 & 0.4467 & 0.3788 \\
 & ACD & 0.7232 & 0.7935 & 0.4289 & 0.3821 & 0.7413 & 0.7538 & 0.4186 & 0.3453 \\
 & \textbf{ARACD} & \textbf{0.7391} & \textbf{0.8078} & \textbf{0.4202} & \textbf{0.3455} & \textbf{0.7490} & \textbf{0.7606} & \textbf{0.4140} & \textbf{0.3341} \\
\midrule
\multirow{3}{*}{NCDM} & Base & 0.6908 & 0.7520 & 0.4527 & 0.4191 & 0.7397 & 0.7451 & 0.4218 & 0.3344 \\
 & ACD & 0.7230 & 0.7938 & 0.4296 & 0.4018 & 0.7421 & 0.7502 & 0.4185 & 0.3336 \\
 & \textbf{ARACD} & \textbf{0.7394} & \textbf{0.8078} & \textbf{0.4203} & \textbf{0.3461} & \textbf{0.7487} & \textbf{0.7617} & \textbf{0.4138} & \textbf{0.3316} \\
\midrule
\multirow{3}{*}{RCD} & Base & 0.7141 & 0.7795 & 0.4354 & 0.3713 & 0.7401 & 0.7477 & 0.4214 & 0.3373 \\
 & ACD & 0.7219 & 0.7915 & 0.4296 & 0.3645 & 0.7445 & 0.7570 & 0.4175 & 0.3381 \\
 & \textbf{ARACD} & \textbf{0.7364} & \textbf{0.8056} & \textbf{0.4210} & \textbf{0.3576} & \textbf{0.7499} & \textbf{0.7611} & \textbf{0.4131} & \textbf{0.3328} \\
\midrule
\multirow{3}{*}{SCD} & Base & 0.7146 & 0.7802 & 0.4350 & 0.3676 & 0.7415 & 0.7549 & 0.4188 & 0.3371 \\
 & ACD & 0.7259 & 0.7930 & 0.4289 & 0.3608 & 0.7460 & 0.7613 & 0.4153 & 0.3305 \\
 & \textbf{ARACD} & \textbf{0.7357} & \textbf{0.8055} & \textbf{0.4212} & \textbf{0.3598} & \textbf{0.7497} & \textbf{0.7618} & \textbf{0.4132} & \textbf{0.3301} \\
\bottomrule
\end{tabular}%
}
\end{table*}

Table 5 highlights that ARACD provides consistently improved response prediction across the two supervised datasets. Compared with Base, ARACD improves ACC and AUC and reduces RMSE and MAE for every reported backbone-dataset pair. Compared with ACD, ARACD yields higher ACC and AUC and lower RMSE and MAE in all 12 comparisons. These patterns suggest that ability-residual modeling can complement supervised affect modeling by accounting for part of the remaining cognitive residual variation. However, the magnitude of the gain still depends on the backbone and dataset.

We further run CACD and ARCACD on ASSIST2009 and Junyi under the same six-backbone setting to supplement cross-dataset evidence. Table 6 reports ACC, AUC, RMSE, and MAE for the two datasets side by side. For ASSIST2009, Base results for each backbone are included as response prediction controls without affective modules or ability residuals.

\begin{table*}[htbp]
\centering
\caption{Response prediction results on ASSIST2009 and Junyi.}\label{tab:6}
\scriptsize
\setlength{\tabcolsep}{2pt}
\renewcommand{\arraystretch}{1.12}
\resizebox{0.95\textwidth}{!}{%
\begin{tabular}{@{}cccccccccc@{}}
\toprule
\multirow{2}{*}{Backbone} & \multirow{2}{*}{Method} & \multicolumn{4}{c}{ASSIST2009} & \multicolumn{4}{c}{Junyi} \\
\cmidrule(lr){3-6} \cmidrule(lr){7-10}
 &  & ACC $\uparrow$ & AUC $\uparrow$ & RMSE $\downarrow$ & MAE $\downarrow$ & ACC $\uparrow$ & AUC $\uparrow$ & RMSE $\downarrow$ & MAE $\downarrow$ \\
\midrule
\multirow{3}{*}{DINA} & Base & 0.7015 & 0.7206 & 0.4590 & 0.4473 & 0.7281 & 0.7115 & 0.4334 & 0.3866 \\
 & CACD & 0.7240 & 0.7513 & 0.4344 & 0.3935 & 0.7456 & 0.7539 & 0.4100 & 0.3575 \\
 & \textbf{ARCACD} & \textbf{0.7786} & \textbf{0.8396} & \textbf{0.3896} & \textbf{0.2799} & \textbf{0.7658} & \textbf{0.7582} & \textbf{0.4013} & \textbf{0.3298} \\
\midrule
\multirow{3}{*}{IRT} & Base & 0.7175 & 0.7354 & 0.4391 & 0.3979 & 0.7335 & 0.7207 & 0.4113 & 0.3554 \\
 & CACD & 0.7255 & 0.7571 & 0.4289 & 0.3581 & 0.7686 & 0.7583 & 0.4009 & 0.3284 \\
 & \textbf{ARCACD} & \textbf{0.7819} & \textbf{0.8428} & \textbf{0.3858} & \textbf{0.2887} & \textbf{0.7692} & \textbf{0.7591} & \textbf{0.4001} & \textbf{0.3251} \\
\midrule
\multirow{3}{*}{MIRT} & Base & 0.6894 & 0.6986 & 0.4574 & 0.4414 & 0.7355 & 0.7243 & 0.4215 & 0.3877 \\
 & CACD & 0.7242 & 0.7499 & 0.4313 & 0.3810 & 0.7580 & 0.7537 & 0.4050 & 0.3436 \\
 & \textbf{ARCACD} & \textbf{0.7807} & \textbf{0.8415} & \textbf{0.3882} & \textbf{0.2786} & \textbf{0.7683} & \textbf{0.7597} & \textbf{0.4004} & \textbf{0.3271} \\
\midrule
\multirow{3}{*}{NCDM} & Base & 0.7188 & 0.7513 & 0.4358 & 0.3921 & 0.7262 & 0.7162 & 0.4453 & 0.4001 \\
 & CACD & 0.7253 & 0.7573 & 0.4332 & 0.3845 & 0.7396 & 0.7277 & 0.4289 & 0.3705 \\
 & \textbf{ARCACD} & \textbf{0.7884} & \textbf{0.8487} & \textbf{0.3823} & \textbf{0.2740} & \textbf{0.7682} & \textbf{0.7596} & \textbf{0.4007} & \textbf{0.3338} \\
\midrule
\multirow{3}{*}{RCD} & Base & 0.7249 & 0.7618 & 0.4271 & 0.3518 & 0.7595 & 0.7589 & 0.4052 & 0.3327 \\
 & CACD & 0.7284 & 0.7584 & 0.4291 & 0.3670 & 0.7706 & 0.7631 & 0.3986 & 0.3174 \\
 & \textbf{ARCACD} & \textbf{0.7787} & \textbf{0.8384} & \textbf{0.3920} & \textbf{0.2747} & \textbf{0.7715} & \textbf{0.7634} & \textbf{0.3971} & \textbf{0.3168} \\
\midrule
\multirow{3}{*}{SCD} & Base & 0.7257 & 0.7621 & 0.4263 & 0.3594 & 0.7570 & 0.7499 & 0.4097 & 0.3572 \\
 & CACD & 0.7273 & 0.7599 & 0.4320 & 0.3358 & 0.7702 & 0.7570 & 0.4045 & 0.3506 \\
 & \textbf{ARCACD} & \textbf{0.7782} & \textbf{0.8387} & \textbf{0.3956} & \textbf{0.2632} & \textbf{0.7704} & \textbf{0.7583} & \textbf{0.4043} & \textbf{0.3501} \\
\bottomrule
\end{tabular}%
}
\end{table*}

Table 6 reveals that on ASSIST2009, ARCACD achieves higher ACC/AUC and lower RMSE/MAE than both Base and CACD on all six backbones. In particular, NCDM+ARCACD reaches ACC/AUC of 0.7884/0.8487 under the targeted configuration with fixed $d_{att}=16,rank=1024$. In supplementary runs that combine ability residuals with the CACD contrastive loss, DINA, IRT, MIRT, RCD, and SCD obtain AUC values of approximately 0.838--0.843. On the Junyi subset, ARCACD also yields consistent improvements on all four response metrics across six backbones. Overall, in the absence of real affect labels, joint modeling of ability residuals and contrastive affective representations yields consistent response-prediction gains across the reported datasets and backbones.

\subsection{Affect Alignment Analysis (RQ2)}

This section answers RQ2, namely whether explicit ability residuals can help the affective branch better align with real affect when real affect labels are available. Table 7 reports affect metrics and presents ASSIST2017 and ASSIST2012 side by side. Since Base does not contain an affective branch, the table includes only ACD and ARACD trained with real affect label supervision. The metrics are Affect RMSE and Affect MAE.

\begin{table*}[htbp]
\centering
\caption{Affect alignment results under real affect labels.}\label{tab:7}
\scriptsize
\setlength{\tabcolsep}{3pt}
\renewcommand{\arraystretch}{1.12}
\resizebox{0.90\textwidth}{!}{%
\begin{tabular}{@{}cccccc@{}}
\toprule
\multirow{2}{*}{Backbone} & \multirow{2}{*}{Method} & \multicolumn{2}{c}{ASSIST2017} & \multicolumn{2}{c}{ASSIST2012} \\
\cmidrule(lr){3-4} \cmidrule(lr){5-6}
 &  & Affect RMSE $\downarrow$ & Affect MAE $\downarrow$ & Affect RMSE $\downarrow$ & Affect MAE $\downarrow$ \\
\midrule
\multirow{2}{*}{DINA} & ACD & 0.2614 & 0.1914 & 0.2184 & 0.1502 \\
 & \textbf{ARACD} & \textbf{0.1990} & \textbf{0.1489} & \textbf{0.1937} & \textbf{0.1239} \\
\midrule
\multirow{2}{*}{IRT} & ACD & 0.2461 & 0.1785 & 0.2453 & 0.1792 \\
 & \textbf{ARACD} & \textbf{0.1925} & \textbf{0.1389} & \textbf{0.1919} & \textbf{0.1179} \\
\midrule
\multirow{2}{*}{MIRT} & ACD & 0.2640 & 0.1900 & 0.2495 & 0.1814 \\
 & \textbf{ARACD} & \textbf{0.1922} & \textbf{0.1390} & \textbf{0.1913} & \textbf{0.1203} \\
\midrule
\multirow{2}{*}{NCDM} & ACD & 0.2266 & 0.1640 & 0.2422 & 0.1703 \\
 & \textbf{ARACD} & \textbf{0.1944} & \textbf{0.1409} & \textbf{0.1922} & \textbf{0.1196} \\
\midrule
\multirow{2}{*}{RCD} & ACD & 0.2055 & \textbf{0.1441} & 0.1966 & 0.1245 \\
 & \textbf{ARACD} & \textbf{0.2038} & 0.1571 & \textbf{0.1942} & \textbf{0.1213} \\
\midrule
\multirow{2}{*}{SCD} & ACD & 0.2054 & \textbf{0.1436} & 0.1996 & 0.1269 \\
 & \textbf{ARACD} & \textbf{0.2037} & 0.1532 & \textbf{0.1945} & \textbf{0.1246} \\
\bottomrule
\end{tabular}%
}
\end{table*}

Table 7 provides generally supportive but non-uniform evidence for improved affect alignment after introducing ability residuals. On ASSIST2012, ARACD obtains lower Affect RMSE and Affect MAE than the corresponding ACD on all six backbones. On ASSIST2017, ARACD lowers both affect metrics for DINA, IRT, MIRT, and NCDM, and reduces Affect RMSE for RCD and SCD, although Affect MAE increases on these two backbones. These findings highlight that ability residuals reduce the pressure on the affective branch to absorb cognitive bias in many settings. Still, affect-alignment gains should not be interpreted as guaranteed for every backbone and metric. This result also indicates that similar response metrics do not necessarily imply affective representation alignment; therefore, later contamination diagnostics separately analyze whether affective representations absorb cognitive residuals.

\subsection{Ablation Study (RQ3)}

This section answers RQ3 by comparing the full model with three variants: Only stu-item residual, Only concept residual, and w/o ability residual. ASSIST2012/ASSIST2017 use supervised ARACD training, whereas ASSIST2009/Junyi use ARCACD training without affect labels. All variants are evaluated under the same data split and model-selection protocol within each dataset.

\begin{table*}[htbp]
\centering
\caption{Hierarchical ablation of supervised ability residual layers on ASSIST2017 and ASSIST2012.}\label{tab:8}
\scriptsize
\setlength{\tabcolsep}{2.5pt}
\renewcommand{\arraystretch}{0.94}
\resizebox{0.98\textwidth}{!}{%
\begin{tabular}{@{}ccccccccccccc@{}}
\toprule
\multirow{2}{*}{Variant} & \multicolumn{6}{c}{ASSIST2017} & \multicolumn{6}{c}{ASSIST2012} \\
\cmidrule(lr){2-7} \cmidrule(lr){8-13}
 & ACC $\uparrow$ & AUC $\uparrow$ & RMSE $\downarrow$ & MAE $\downarrow$ & Affect RMSE $\downarrow$ & Affect MAE $\downarrow$ & ACC $\uparrow$ & AUC $\uparrow$ & RMSE $\downarrow$ & MAE $\downarrow$ & Affect RMSE $\downarrow$ & Affect MAE $\downarrow$ \\
\midrule
\textbf{Full ARACD} & \textbf{0.7394} & \textbf{0.8078} & \textbf{0.4203} & \textbf{0.3461} & \textbf{0.1944} & \textbf{0.1409} & \textbf{0.7487} & \textbf{0.7617} & \textbf{0.4138} & \textbf{0.3316} & \textbf{0.1922} & \textbf{0.1196} \\
Only stu-item residual & 0.7385 & 0.8051 & 0.4219 & 0.3494 & 0.1988 & 0.1456 & 0.7444 & 0.7537 & 0.4170 & 0.3328 & 0.1942 & 0.1210 \\
Only concept residual & 0.7288 & 0.7994 & 0.4286 & 0.3839 & 0.2082 & 0.1484 & 0.7457 & 0.7526 & 0.4156 & 0.3321 & 0.1973 & 0.1258 \\
w/o ability residual & 0.7230 & 0.7938 & 0.4296 & 0.4018 & 0.2266 & 0.1640 & 0.7421 & 0.7502 & 0.4185 & 0.3336 & 0.2422 & 0.1703 \\
\bottomrule
\end{tabular}%
}
\end{table*}

Table 8 reveals that in the supervised setting, the complete model achieves the best result on every reported response and affect metric across ASSIST2017 and ASSIST2012. The student-item-only and concept-only variants are consistently weaker than the complete model, indicating that the two residual layers provide complementary information.

\begin{table*}[htbp]
\centering
\caption{Hierarchical ablation of ability residual layers on ASSIST2009 and Junyi.}\label{tab:9}
\scriptsize
\setlength{\tabcolsep}{3pt}
\renewcommand{\arraystretch}{0.94}
\resizebox{0.90\textwidth}{!}{%
\begin{tabular}{@{}ccccccccc@{}}
\toprule
\multirow{2}{*}{Variant} & \multicolumn{4}{c}{ASSIST2009} & \multicolumn{4}{c}{Junyi} \\
\cmidrule(lr){2-5} \cmidrule(lr){6-9}
 & ACC $\uparrow$ & AUC $\uparrow$ & RMSE $\downarrow$ & MAE $\downarrow$ & ACC $\uparrow$ & AUC $\uparrow$ & RMSE $\downarrow$ & MAE $\downarrow$ \\
\midrule
\textbf{Full ARCACD} & \textbf{0.7884} & \textbf{0.8487} & \textbf{0.3823} & \textbf{0.2740} & \textbf{0.7682} & \textbf{0.7596} & \textbf{0.4007} & \textbf{0.3232} \\
Only stu-item residual & 0.7804 & 0.8388 & 0.3872 & 0.2909 & 0.7599 & 0.7517 & 0.4045 & 0.3354 \\
Only concept residual & 0.7333 & 0.7872 & 0.4116 & 0.3388 & 0.7636 & 0.7478 & 0.4032 & 0.3338 \\
w/o ability residual & 0.7253 & 0.7573 & 0.4332 & 0.3845 & 0.7396 & 0.7277 & 0.4289 & 0.3705 \\
\bottomrule
\end{tabular}%
}
\end{table*}

Accordingly, Table 9 reveals a similar pattern in the setting without affect labels. On both ASSIST2009 and Junyi, the full model achieves the best ACC/AUC and the lowest RMSE/MAE among all variants. The student-item layer explains most of the gain, while the concept layer provides complementary concept-related corrections when combined with it.

\subsection{Hyperparameter Analysis (RQ4)}

This section answers RQ4 by examining two capacity parameters on NCDM: the attention dimension $d_{att}$ and the student-item low-rank dimension $rank$. ASSIST2017 uses supervised affect prediction, while ASSIST2009 uses contrastive ARCACD without affect labels. Figures 3 and 4 report the complete sensitivity comparisons: Figure 3 uses fixed $rank=1024$ and sweeps $d_{att}=4,8,16,32$, while Figure 4 uses fixed $d_{att}=16$ and sweeps $rank$ from 4 to 2048.

\begin{figure}[htbp]
\centering
\begin{minipage}[t]{0.48\linewidth}
\centering
\includegraphics[width=\linewidth]{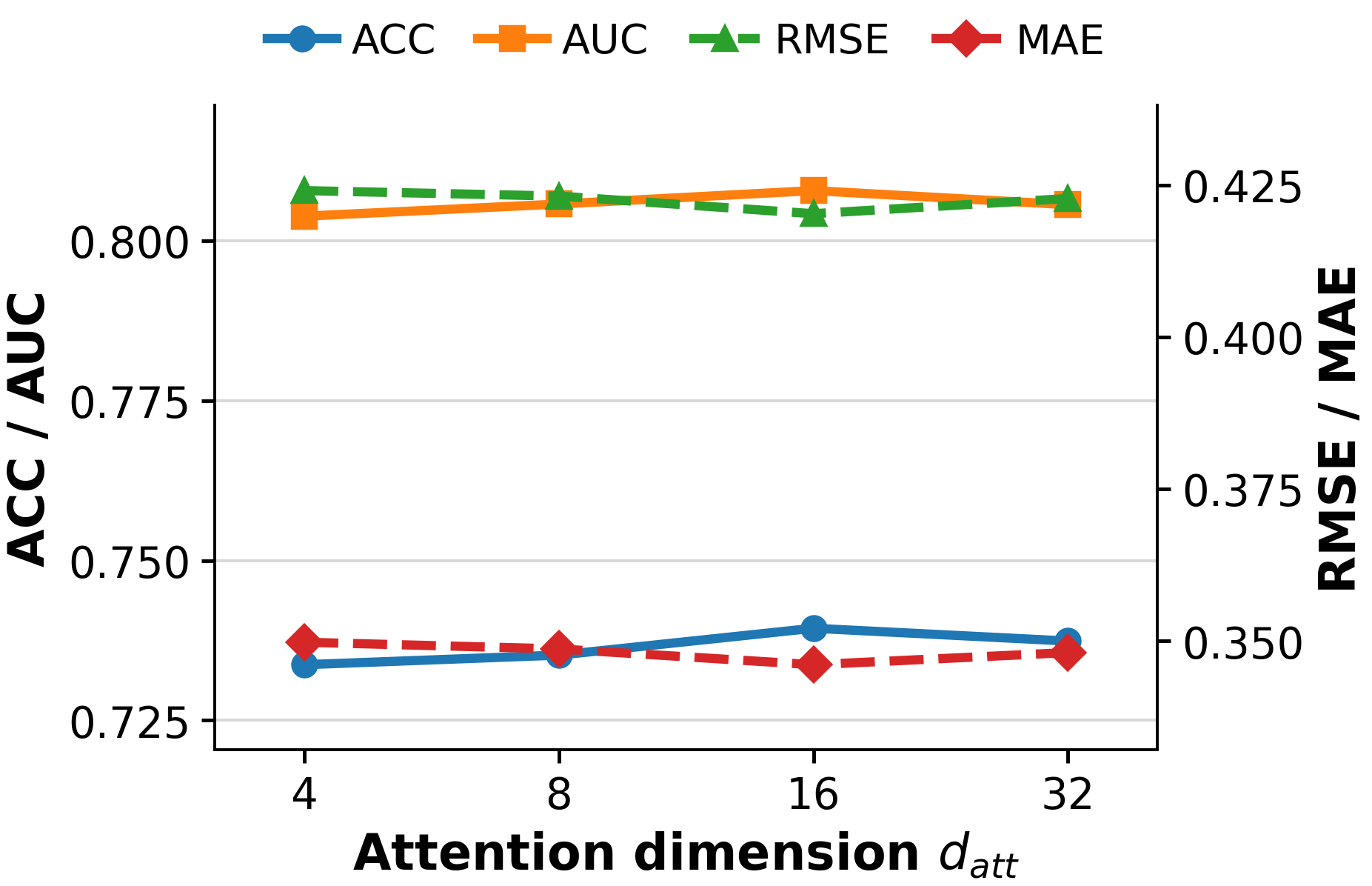}
\small (a) ASSIST2017 response
\end{minipage}
\hfill
\begin{minipage}[t]{0.48\linewidth}
\centering
\includegraphics[width=\linewidth]{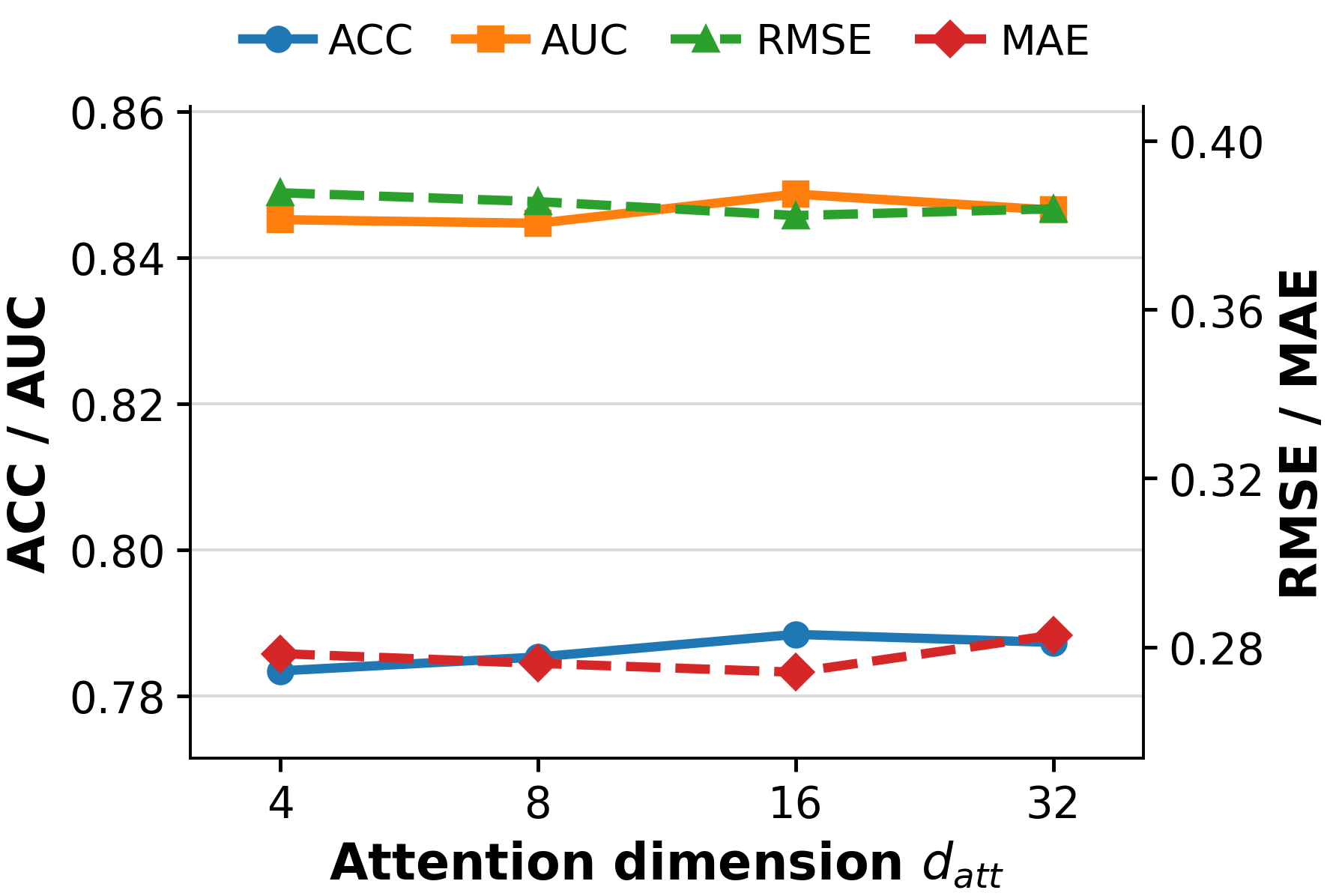}
\small (b) ASSIST2009 response
\end{minipage}
\hfill
\begin{minipage}[t]{0.62\linewidth}
\centering
\includegraphics[width=\linewidth]{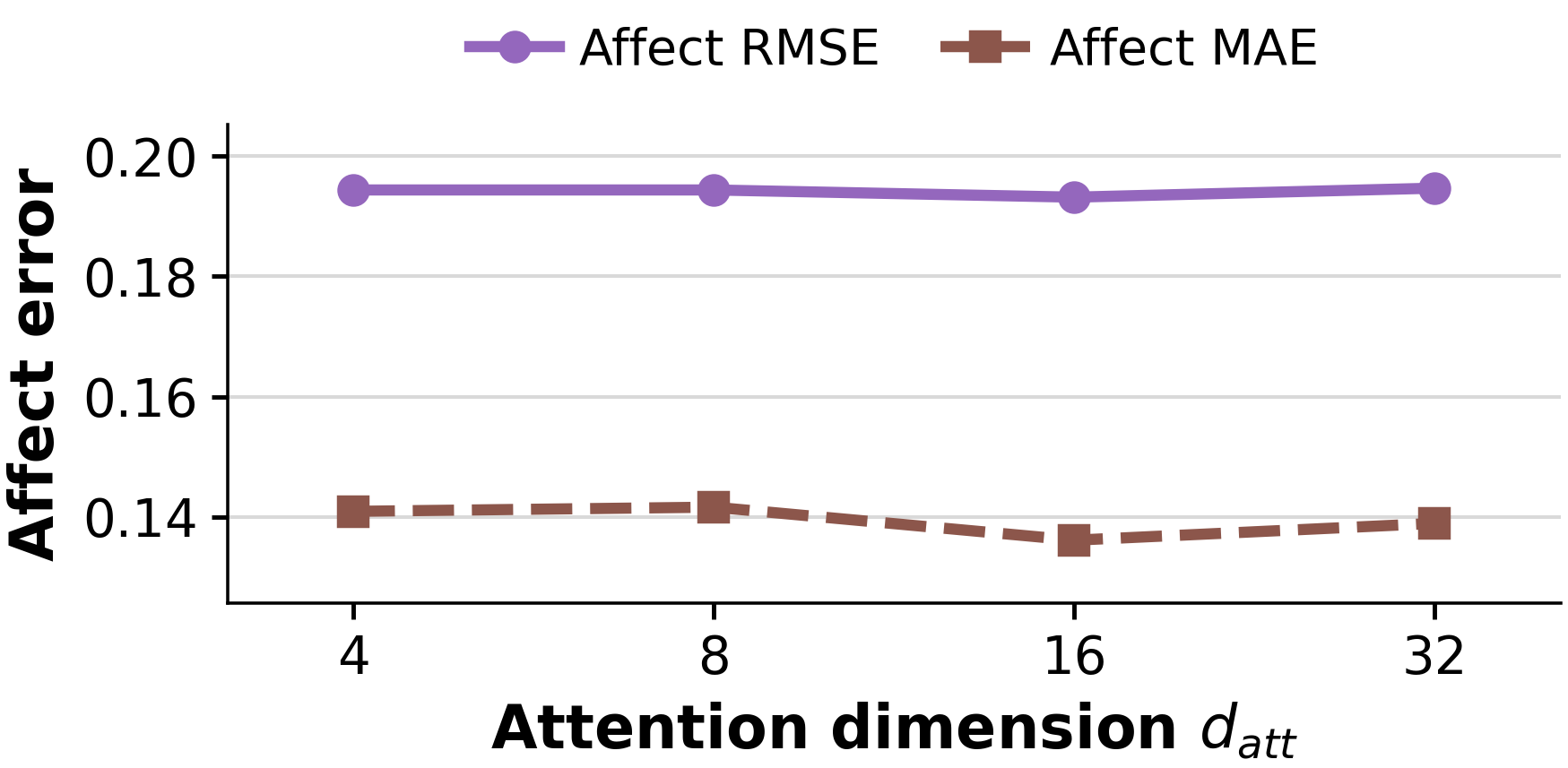}
\small (c) ASSIST2017 affect
\end{minipage}
\caption{Sensitivity to the concept-attention dimension $d_{att}$ with fixed $rank=1024$. Panels (a) and (b) report ACC/AUC and RMSE/MAE; panel (c) reports Affect RMSE/MAE.}
\label{fig:3}
\end{figure}

Figure 3 demonstrates that $d_{att}$ has a limited but consistent effect once $rank=1024$ is fixed. Both datasets obtain their best or near-best response metrics around $d_{att}=16$, while increasing it to 32 introduces no stable gain. Thus, concept residual attention only needs moderate capacity when the student-item residual layer is sufficiently expressive.

\begin{figure}[htbp]
\centering
\begin{minipage}[t]{0.48\linewidth}
\centering
\includegraphics[width=\linewidth]{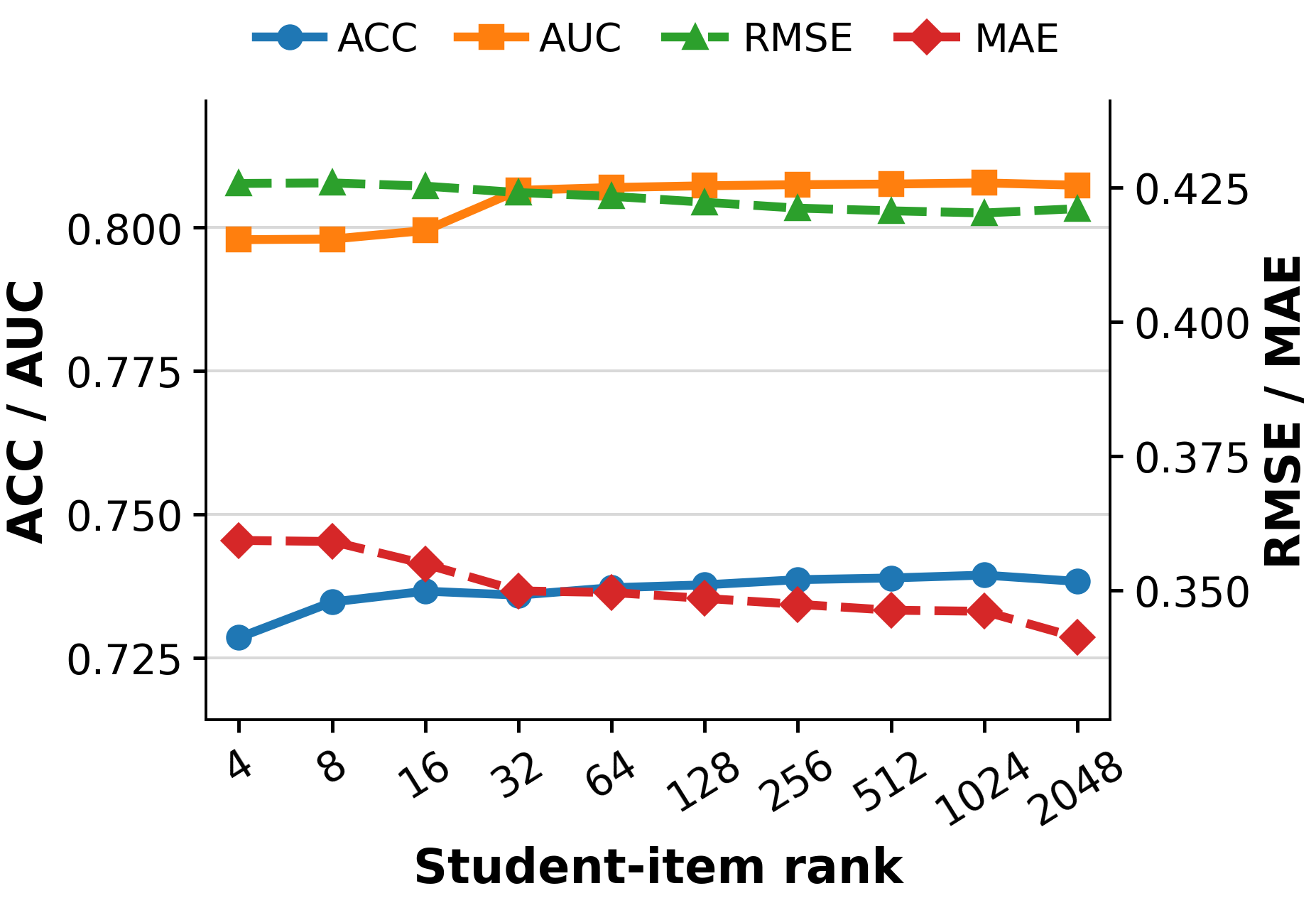}
\small (a) ASSIST2017 response
\end{minipage}
\hfill
\begin{minipage}[t]{0.48\linewidth}
\centering
\includegraphics[width=\linewidth]{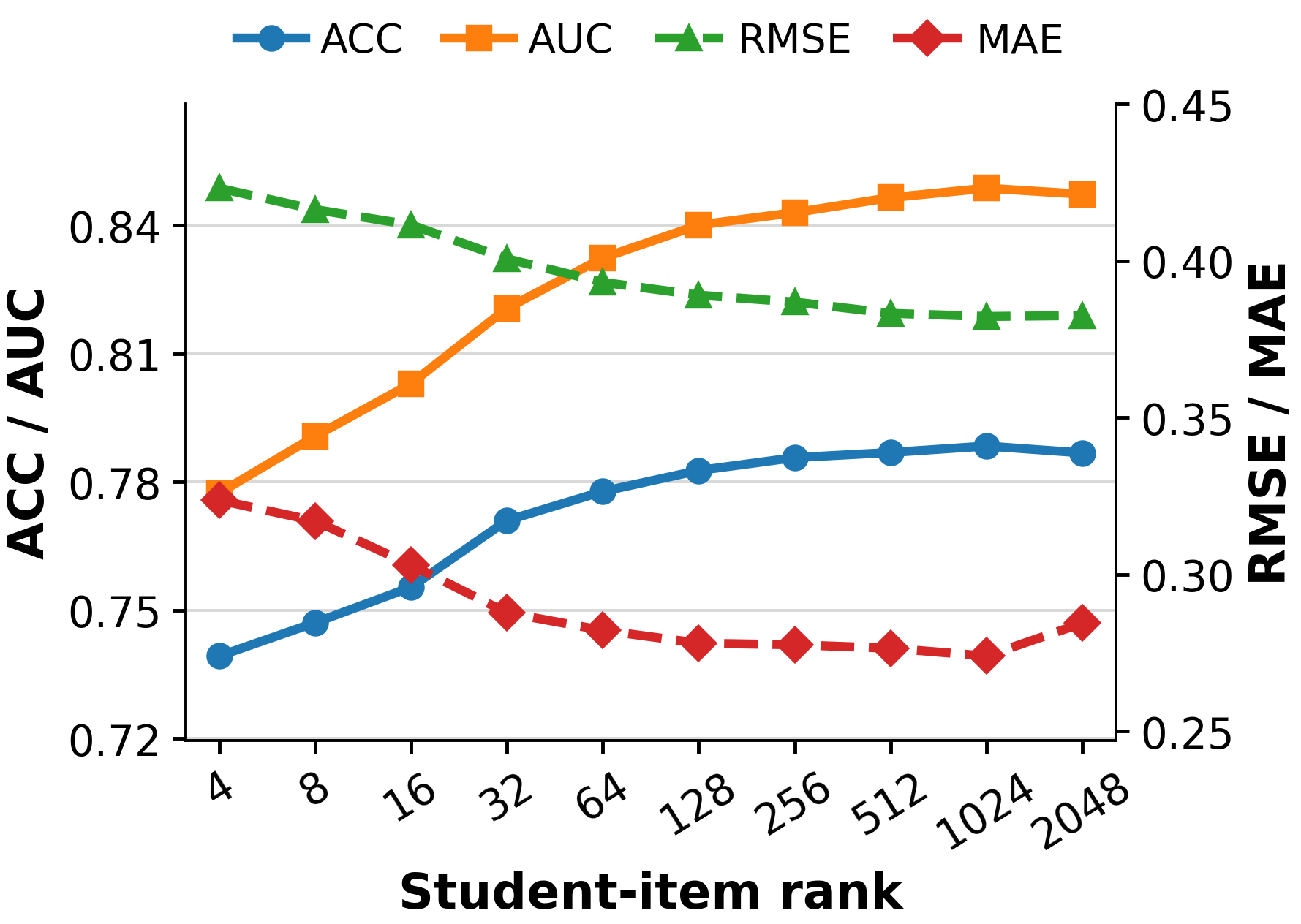}
\small (b) ASSIST2009 response
\end{minipage}
\hfill
\begin{minipage}[t]{0.62\linewidth}
\centering
\includegraphics[width=\linewidth]{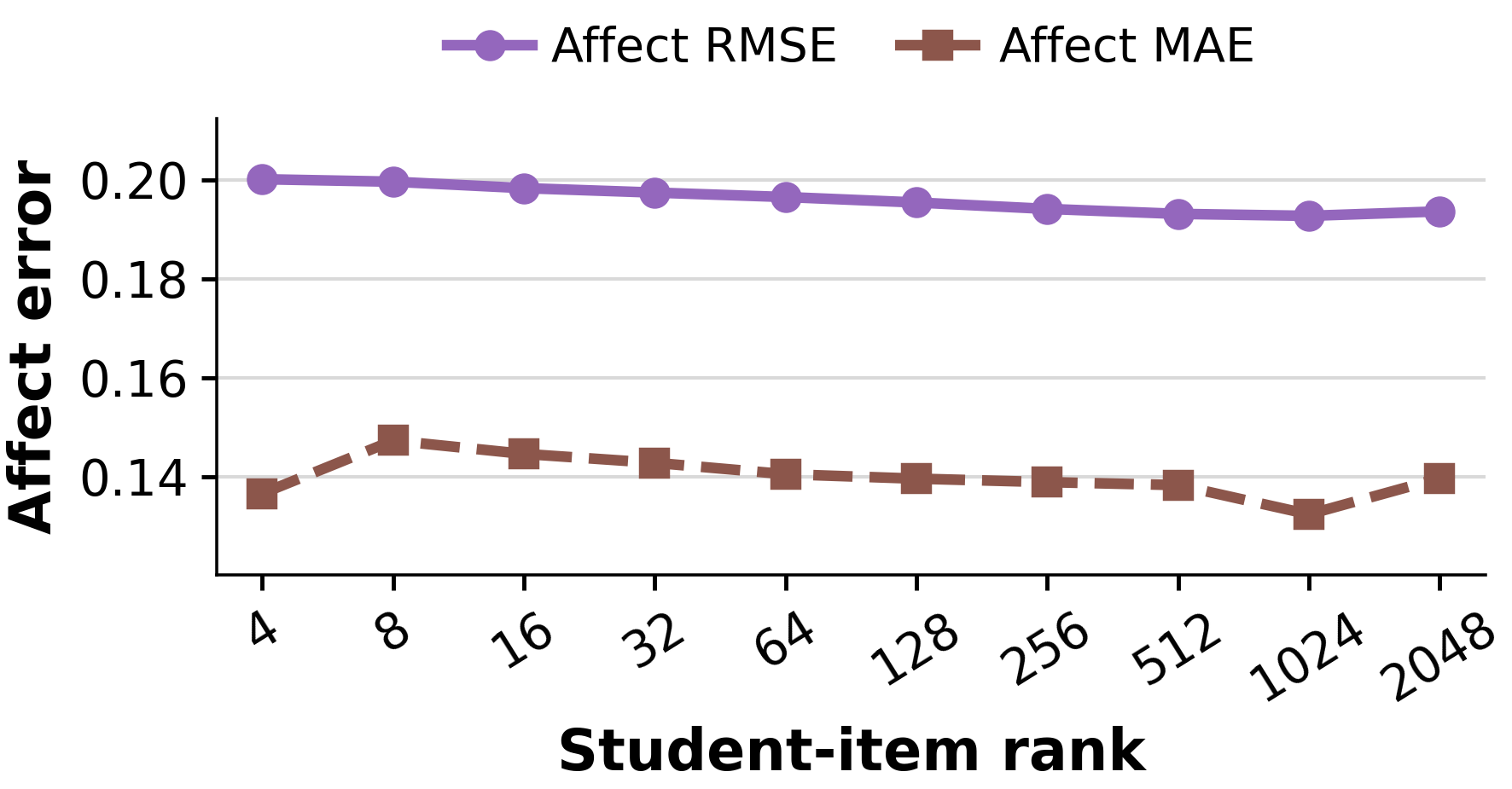}
\small (c) ASSIST2017 affect
\end{minipage}
\caption{Sensitivity to the student-item low-rank dimension $rank$ with fixed $d_{att}=16$. Panels (a) and (b) report ACC/AUC and RMSE/MAE; panel (c) reports Affect RMSE/MAE.}
\label{fig:4}
\end{figure}

Figure 4 highlights that $rank$ is the more important capacity source. Increasing $rank$ from 4 to 1024 markedly improves ASSIST2009 response metrics and mildly improves ASSIST2017, while $rank=2048$ yields limited additional benefit and can increase some errors. Affect prediction on ASSIST2017 follows the same pattern, with lower error around $rank=1024$. Overall, $rank=1024$ provides a practical balance between response accuracy, affect alignment, and capacity saturation.

\subsection{Residual Map Analysis (RQ5)}

This section answers RQ5 by visualizing the residual space using $r^{stu\text{-}item}_{si}$ and $r^{concept}_{si}$ as two-dimensional coordinates. Figure 5 presents the sampled test interactions from all four datasets, colored by true response labels. The residual spaces exhibit response-related directions rather than random perturbations: ASSIST2017 and ASSIST2009 are dominated by the student-item residual direction, ASSIST2012 shows a smoother but still visible offset, and Junyi uses both residual types. This provides representation-level evidence that the two residual layers absorb complementary cognitive bias.

\begin{figure}[htbp]
\centering
\includegraphics[width=0.96\linewidth]{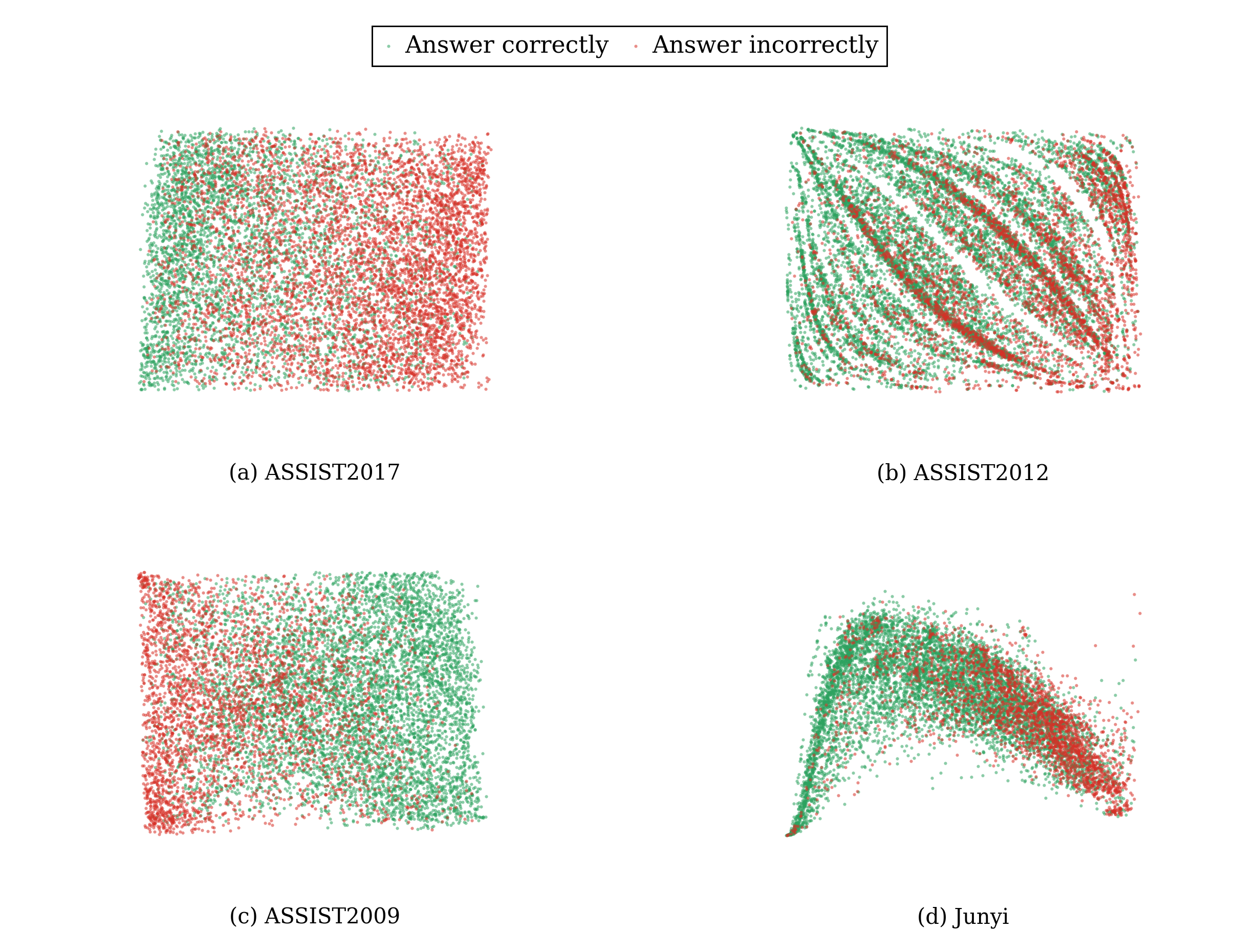}
\caption{Ability-residual interaction maps for ASSIST2017, ASSIST2012, ASSIST2009, and Junyi.}
\label{fig:5}
\end{figure}

\subsection{Affect Contamination Diagnostics (RQ6)}

This section answers RQ6 by testing whether ability residuals reduce cognitive leakage into affective representations. Since response metrics alone cannot determine whether the affective branch has learned a semantically clean non-cognitive state, we construct a synthetic contamination probe. A known bias term is added to the response-generation process, composed of student, item, and concept effects, and ARACD is trained with $d_{att}=16$ and $rank=1024$ without exposing the bias term. Then we test whether this term is easier to recover from the affective branch or the ability residual branch. Three diagnostics are reported: mean abs corr$(affect,bias)$, affect probe $R^2$, and mean abs corr(ability residual,bias).

\begin{center}
\begin{minipage}{\linewidth}
\centering
\refstepcounter{table}\label{tab:10}
{\small\textbf{Table \thetable}}\par
{\small Affect contamination leakage diagnostics on ASSIST2017.}\par\vspace{2pt}
\tiny
\setlength{\tabcolsep}{1.5pt}
\renewcommand{\arraystretch}{1.12}
\resizebox{\linewidth}{!}{%
\begin{tabular}{@{}cccccc@{}}
\toprule
Variant & \makecell{Affect RMSE\\$\downarrow$} & \makecell{Affect MAE\\$\downarrow$} & \makecell{Mean abs corr\\(affect,bias) $\downarrow$} & \makecell{Affect probe\\$R^2$ $\downarrow$} & \makecell{Mean abs corr\\(residual,bias) $\uparrow$} \\
\midrule
Full & 0.1935 & 0.1383 & 0.0665 & 0.0568 & 0.6890 \\
w/o A residual & 0.2298 & 0.1664 & 0.5878 & 0.4716 & n/a \\
\bottomrule
\end{tabular}%
}
\end{minipage}
\end{center}

Table 10 and Figure 6 provide evidence consistent with a redistribution of synthetic cognitive bias away from the affective branch. Without ability residuals, the mean abs corr$(affect,bias)$ and affect probe $R^2$ reach 0.5878 and 0.4716, respectively. With ability residuals, they drop to 0.0665 and 0.0568, while residual-bias correlation reaches 0.6890, and Affect RMSE/MAE also decrease. These diagnostics combined suggest that more of the synthetic contamination signal is represented in the ability-residual pathway rather than in the affective representation.

\begin{figure}[htbp]
\centering
\includegraphics[width=0.82\linewidth]{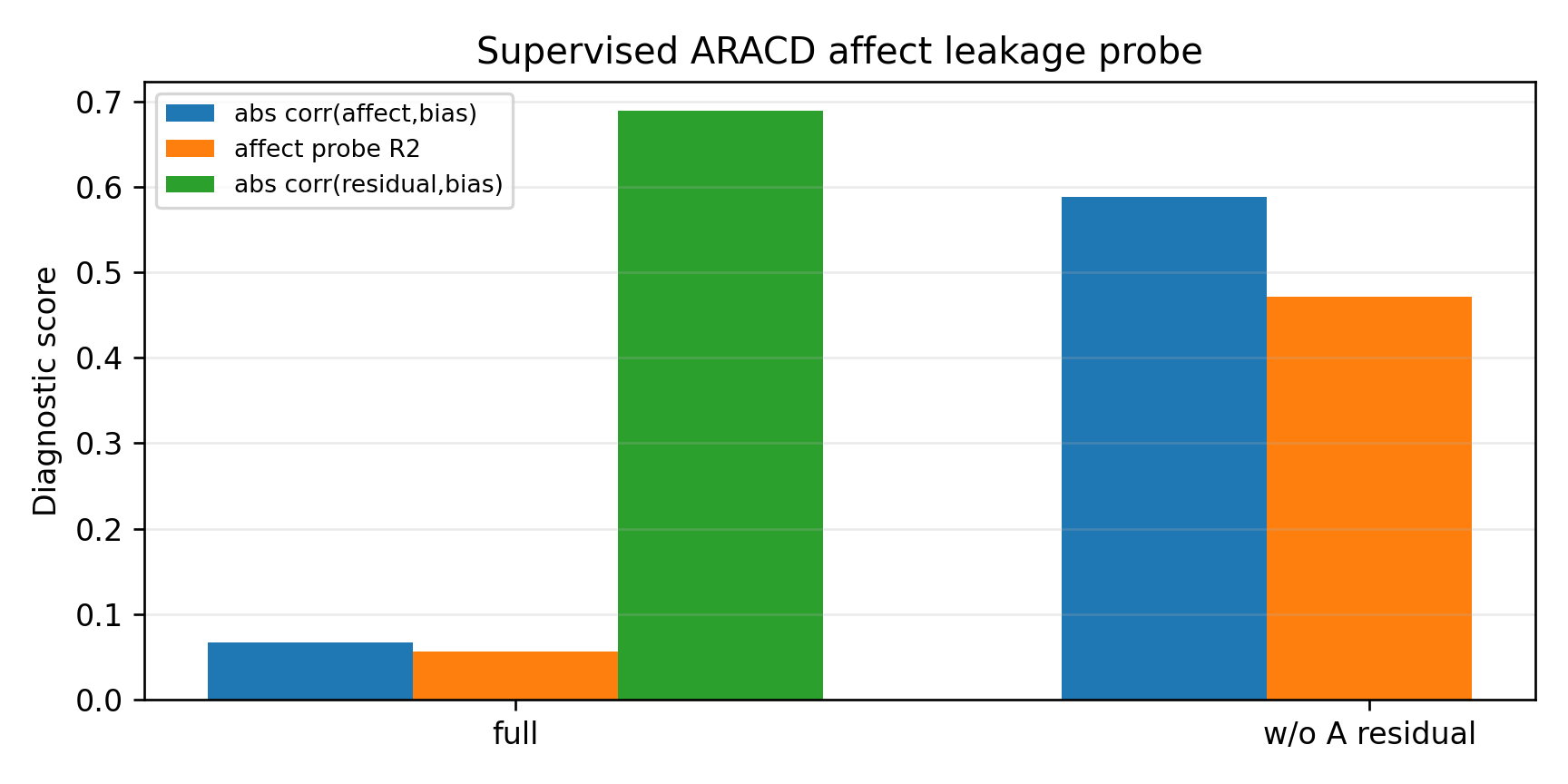}
\caption{Contamination leakage probe of the affective branch.}
\label{fig:6}
\end{figure}

As a complementary view, we project supervised ARACD affect representations from ASSIST2017 into a PCA space and color them by response label and ability residual magnitude. Ideally, the affective representation should preserve response-relevant non-cognitive variation needed for guess/slip modulation, while not being primarily structured by the magnitude of ability residuals.

\begin{figure}[htbp]
\centering
\includegraphics[width=0.94\linewidth]{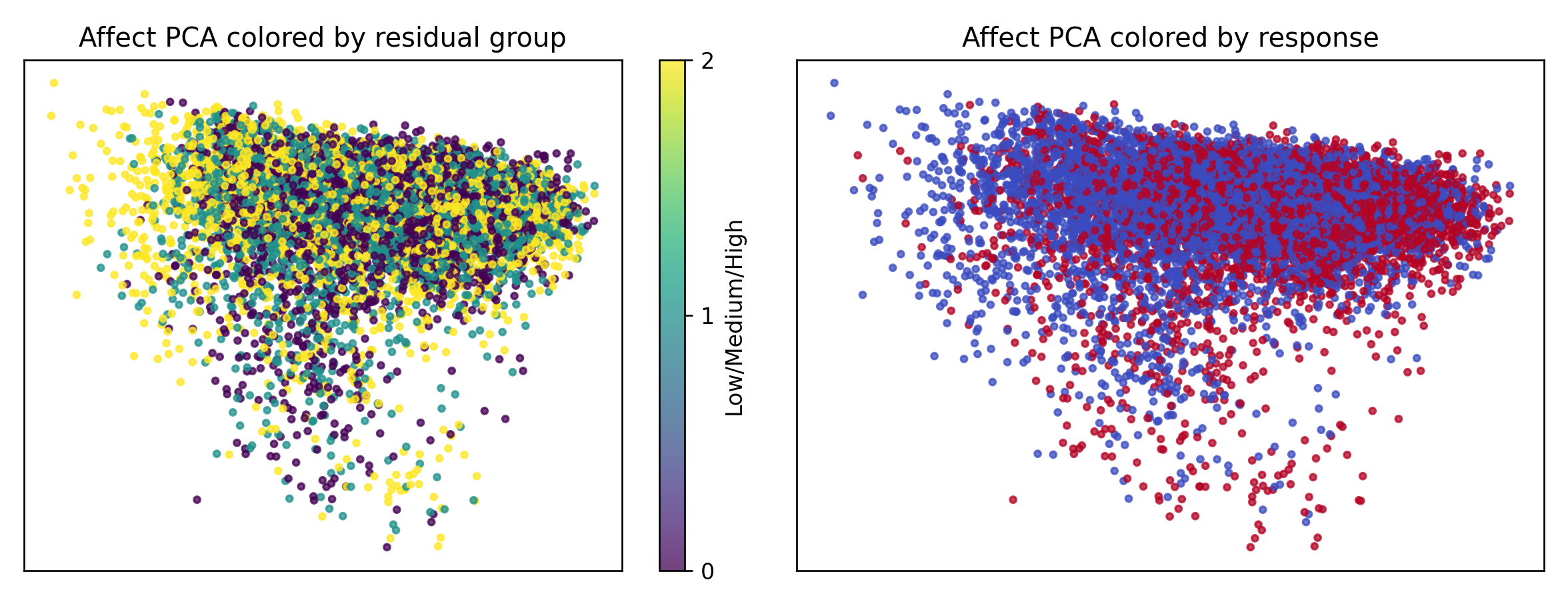}
\caption{PCA visualization of supervised ARACD affective representations on ASSIST2017, colored by ability-residual group (left) and response label (right).}
\label{fig:7}
\end{figure}

\begin{table*}[!t]
\centering
\caption{Absolute correlations between supervised ARACD affect dimensions and residuals/response labels.}\label{tab:11}
\scriptsize
\setlength{\tabcolsep}{4pt}
\renewcommand{\arraystretch}{1.16}
\resizebox{0.92\columnwidth}{!}{%
\begin{tabular}{@{}ccc@{}}
\toprule
Affect dim & \makecell{Abs corr. with\\abs residual $\downarrow$} & \makecell{Abs corr. with\\response $\uparrow$} \\
\midrule
0 & 0.1167 & 0.2114 \\
1 & 0.0958 & 0.2081 \\
2 & 0.0785 & 0.1933 \\
3 & 0.0052 & 0.0011 \\
\bottomrule
\end{tabular}%
}
\end{table*}

Figure 7 and Table 11 support this expectation, as the affect dimensions correlate only weakly with absolute ability residuals (at most 0.1167), while the first three dimensions show clearer response-label correlations. This does not imply a fully isolated affect space, since affect still participates in response prediction. Still, it indicates that ability residuals reduce the tendency of the affect branch to encode stable cognitive bias.

\subsection{Long-Tail Student Results (RQ7)}

This section answers RQ7 by evaluating low-activity students, defined by the lower tertile of training interactions. These students are challenging cases because sparse records provide limited evidence for estimating mastery states, affective representations, and personalized residuals. In this scenario, we compare ACD/\allowbreak RCD/\allowbreak SCD/\allowbreak ARACD on ASSIST2017 and CACD/\allowbreak RCD/\allowbreak SCD/\allowbreak ARCACD on ASSIST2009, using ACC, AUC, RMSE, ECE, and Brier. ECE measures the confidence-accuracy gap over probability bins \cite{ref35}, while Brier score measures the mean squared error between predicted probability and the binary label \cite{ref36}; lower values are better for both.

\begin{center}
\begin{minipage}{\linewidth}
\centering
\refstepcounter{table}\label{tab:12}
{\small\textbf{Table \thetable}}\par
{\small Results for long-tail students.}\par\vspace{2pt}
\scriptsize
\setlength{\tabcolsep}{3pt}
\renewcommand{\arraystretch}{1.12}
\resizebox{\linewidth}{!}{%
\begin{tabular}{@{}ccccccc@{}}
\toprule
Dataset & Method & Low-activity ACC $\uparrow$ & Low-activity AUC $\uparrow$ & Low-activity RMSE $\downarrow$ & Low-activity ECE $\downarrow$ & Low-activity Brier $\downarrow$ \\
\midrule
ASSIST2017 & ACD & 0.6915 & 0.7442 & 0.4547 & 0.0528 & 0.2067 \\
ASSIST2017 & RCD & 0.6922 & 0.7478 & 0.4479 & 0.0263 & 0.2007 \\
ASSIST2017 & SCD & 0.6887 & 0.7466 & 0.4481 & 0.0209 & 0.2008 \\
\textbf{ASSIST2017} & \textbf{ARACD} & \textbf{0.7001} & \textbf{0.7650} & \textbf{0.4405} & \textbf{0.0175} & \textbf{0.1941} \\
ASSIST2009 & CACD & 0.6550 & 0.7365 & 0.4596 & 0.0772 & 0.2112 \\
ASSIST2009 & RCD & 0.7072 & 0.7497 & 0.4404 & \textbf{0.0303} & 0.1939 \\
ASSIST2009 & SCD & 0.7095 & 0.7489 & 0.4413 & 0.0394 & 0.1948 \\
\textbf{ASSIST2009} & \textbf{ARCACD} & \textbf{0.7097} & \textbf{0.7533} & \textbf{0.4398} & 0.0381 & \textbf{0.1934} \\
\bottomrule
\end{tabular}%
}
\end{minipage}
\end{center}

Table 12 reveals that ARACD performs best on all reported low-activity metrics in ASSIST2017. On ASSIST2009, ARCACD achieves the highest AUC and the lowest RMSE/Brier, while ACC and ECE remain close to strong RCD/SCD baselines. These results indicate that ability residuals are especially helpful for ranking and probability-error reduction in sparse student groups.

\subsection{Case Study}

We further analyze student S453 from ASSIST2017 to illustrate how ability residuals correct individual response bias. This student has 388 training interactions and 109 test interactions. After retraining the independent Base NCDM used in the case analysis, Base NCDM obtains ACC/MAE/AUC of 0.6606/0.4371/0.7361 on this student. Notably, ACD still leaves several systematic probability errors for this student, with ACC/MAE/AUC of 0.6330/0.4729/0.7031, while ARACD raises ACC to 0.7523, MAE reduces to 0.3529, and AUC improves to 0.7571, correcting 16 ACD errors while introducing only 3 new errors. This suggests that ability residuals provide directional personalized calibration rather than uniform probability shifts.

\begin{figure}[htbp]
\centering
\begin{minipage}[t]{0.84\linewidth}
\centering
\includegraphics[width=\linewidth]{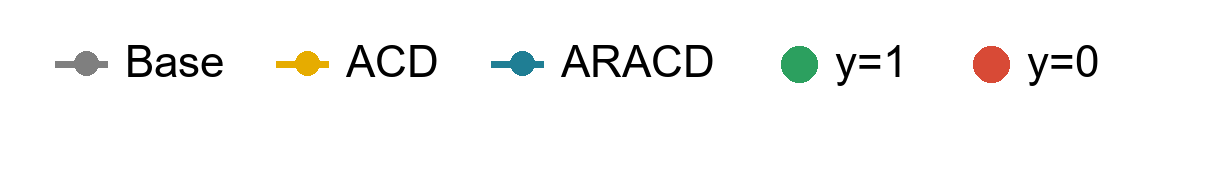}
\small Legend
\end{minipage}
\par\medskip
\begin{minipage}[t]{0.32\linewidth}
\centering
\includegraphics[width=\linewidth]{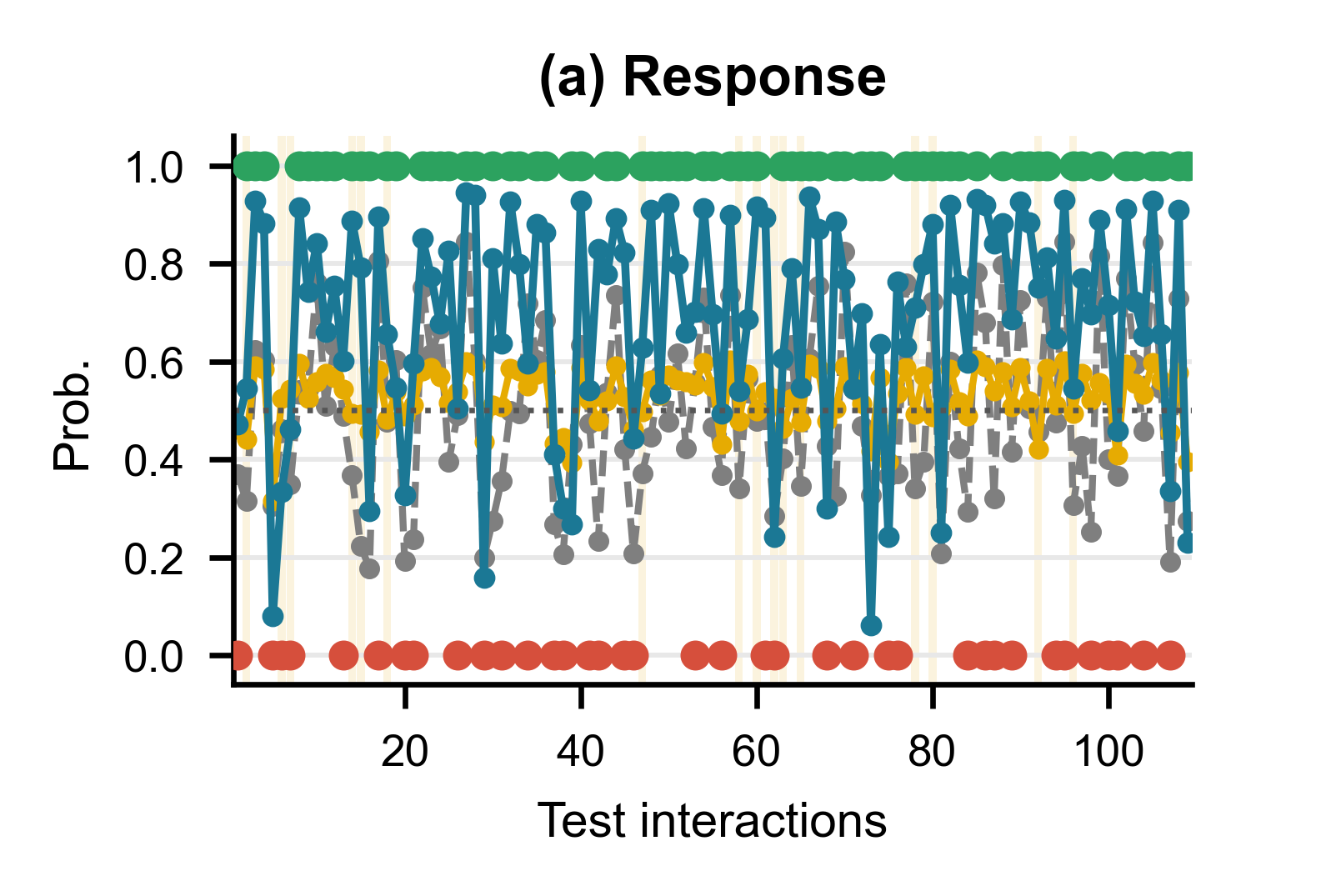}
\small (a) Probability trajectories
\end{minipage}
\hfill
\begin{minipage}[t]{0.32\linewidth}
\centering
\includegraphics[width=\linewidth]{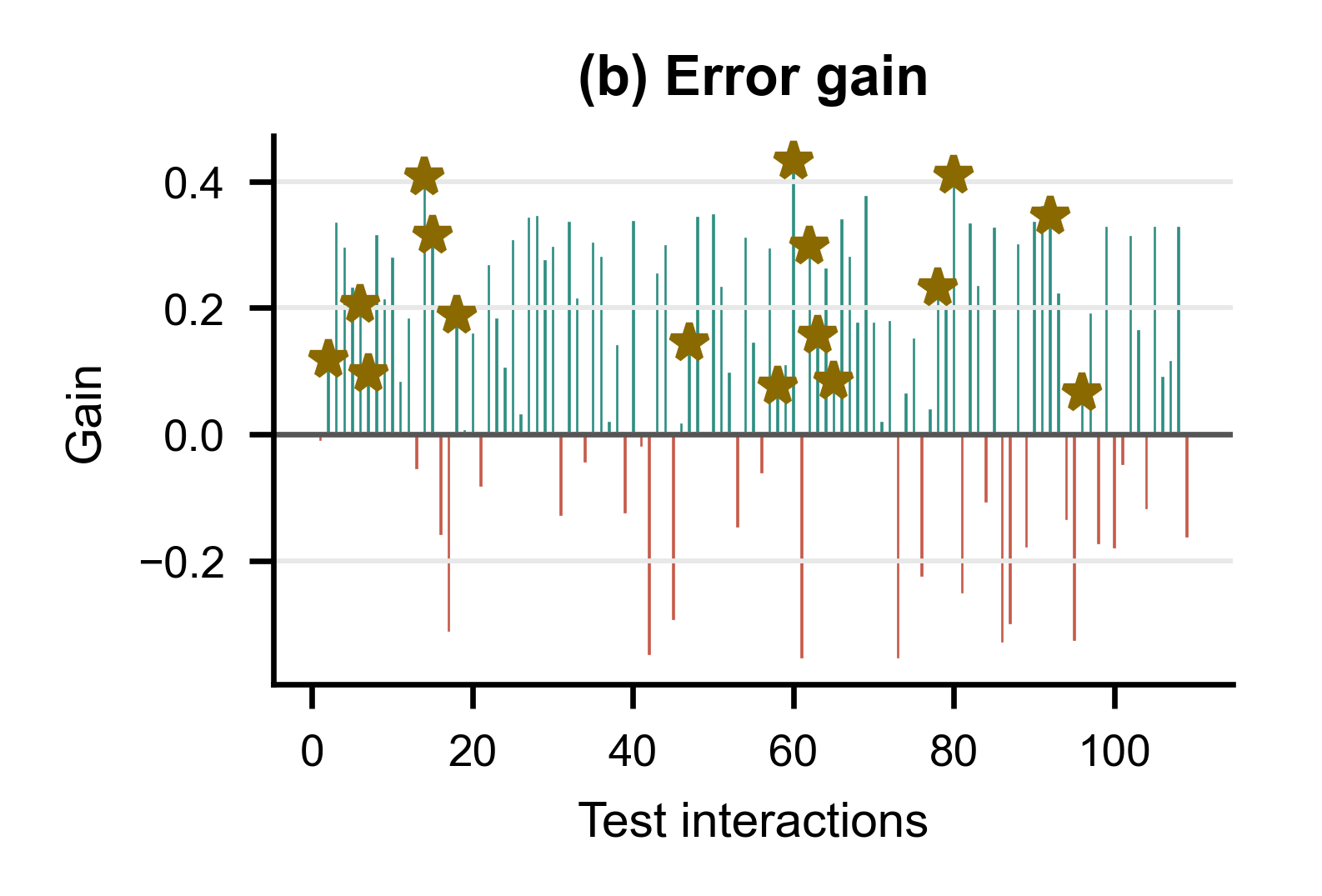}
\small (b) ARACD error reduction over ACD
\end{minipage}
\hfill
\begin{minipage}[t]{0.32\linewidth}
\centering
\includegraphics[width=\linewidth]{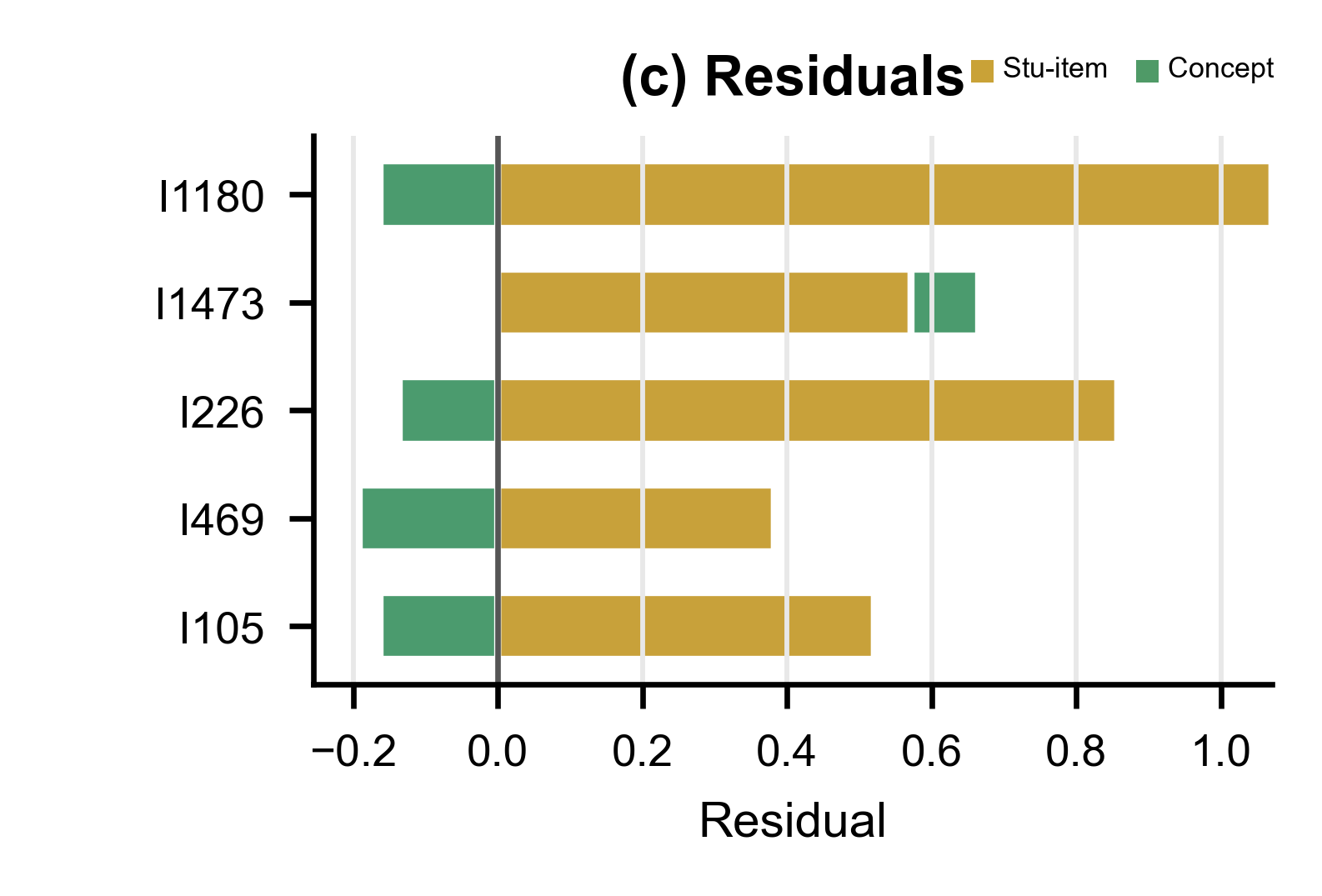}
\small (c) Residual decomposition
\end{minipage}
\caption{Student-level case study for S453. Panels (a)--(c) show probability trajectories, ARACD error reduction over ACD, and residual decomposition.}
\label{fig:8}
\end{figure}

\begin{center}
\begin{minipage}{\linewidth}
\centering
\refstepcounter{table}\label{tab:13}
{\small\textbf{Table \thetable}}\par
{\small Representative interactions and residual decomposition for student S453.}\par\vspace{2pt}
\tiny
\setlength{\tabcolsep}{2pt}
\renewcommand{\arraystretch}{1.12}
\resizebox{\linewidth}{!}{%
\begin{tabular}{@{}ccccccccc@{}}
\toprule
Attempt & Item & Concepts & $y$ & Base prob & ACD prob & ARACD prob & $r^{stu\text{-}item}$ & $r^{concept}$ \\
\midrule
60 & 1180 & 38 & 1 & 0.4787 & 0.4996 & 0.9167 & 1.0712 & -0.1631 \\
80 & 1473 & 15 & 1 & 0.7224 & 0.4862 & 0.8808 & 0.5718 & 0.0934 \\
14 & 226 & 35 & 1 & 0.3678 & 0.4945 & 0.8876 & 0.8574 & -0.1367 \\
62 & 135 & 46 & 0 & 0.2850 & 0.5248 & 0.2414 & -0.3903 & -0.1380 \\
29 & 40 & 14 & 0 & 0.1993 & 0.4367 & 0.1582 & -0.6087 & -0.2034 \\
6 & 1171 & 38 & 0 & 0.4621 & 0.5244 & 0.3337 & -0.2446 & -0.1631 \\
\bottomrule
\end{tabular}%
}
\end{minipage}
\end{center}

Table 13 presents representative corrected interactions. For $y=1$, positive student-item residuals raise ACD probabilities near 0.49 to 0.8808--0.9167; for $y=0$, negative residuals lower probabilities to 0.1582--0.3337. The opposite residual directions across labels show that ARACD performs bidirectional student-item calibration.

\begin{figure}[htbp]
\centering
\includegraphics[width=0.88\linewidth]{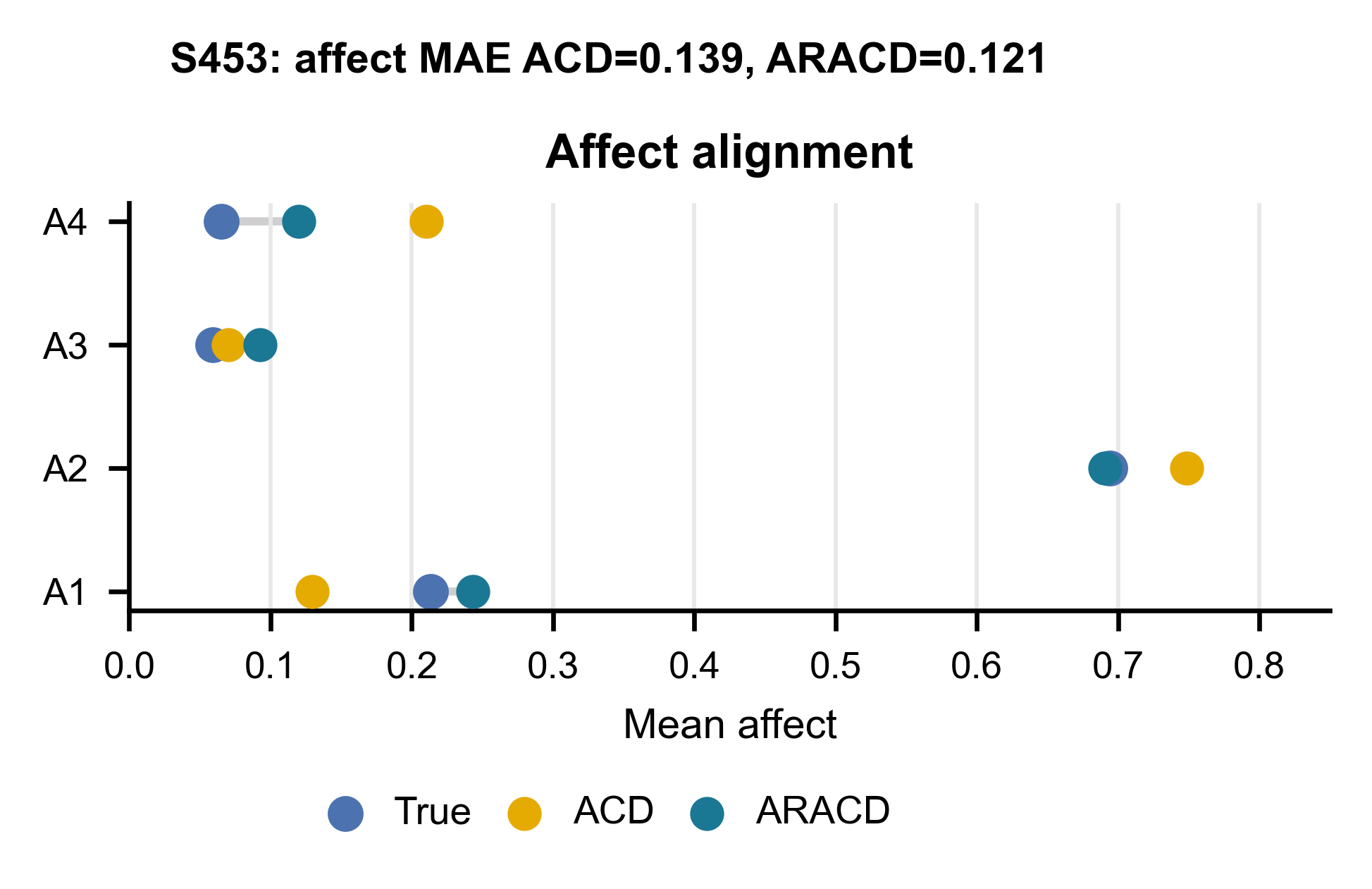}
\caption{Student-level affect analysis for S453. Mean values of the four observed affect dimensions are compared with ACD and ARACD predictions.}
\label{fig:9}
\end{figure}

\begin{center}
\begin{minipage}{\linewidth}
\centering
\refstepcounter{table}\label{tab:14}
{\small\textbf{Table \thetable}}\par
{\small Affect alignment statistics for student S453.}\par\vspace{2pt}
\scriptsize
\setlength{\tabcolsep}{3pt}
\renewcommand{\arraystretch}{1.16}
\resizebox{0.98\linewidth}{!}{%
\begin{tabular}{@{}cccccc@{}}
\toprule
Affect dim & Mean true & ACD pred & ARACD pred & ACD MAE $\downarrow$ & ARACD MAE $\downarrow$ \\
\midrule
A1 & 0.2133 & 0.1293 & 0.2430 & 0.1107 & 0.0966 \\
A2 & 0.6941 & 0.7486 & 0.6906 & 0.1138 & 0.1092 \\
A3 & 0.0591 & 0.0701 & 0.0924 & 0.1083 & 0.1295 \\
A4 & 0.0651 & 0.2101 & 0.1202 & 0.2251 & 0.1507 \\
\bottomrule
\end{tabular}%
}
\end{minipage}
\end{center}

Figure 9 and Table 14 further demonstrate that ARACD preserves affect alignment for S453: overall affect MAE decreases from 0.1395 to 0.1215, with improvements on A1, A2, and A4. Thus, the response improvement does not come at the expense of affect prediction; ability residuals absorb stable student-item offsets while reducing pressure on the affective representation. Overall, this case study shows that ability residuals can correct individualized response bias while keeping affect predictions aligned with external labels.

\section{Conclusion}

This paper studies the affect contamination problem in affective cognitive diagnosis. Unlike approaches that treat the affective module mainly as a compensator for backbone prediction errors, this study focuses on the base cognitive diagnosis backbone, which may still leave stable cognitive bias unmodeled. Indeed, if such bias is not explicitly modeled, affective representations may absorb information about items, concepts, student ability, or student-item matching, making them less specific to non-cognitive states. Motivated by this observation, this work introduces an ability-residual decoupling framework that assigns stable cognitive residuals first to the student-item residual layer and the concept residual layer, and then to the affective module model guess/slip modulation. This provides a structured separation between cognitive residuals and non-cognitive factors.

Experimental results support the proposed approach from multiple perspectives. Specifically, the main experiments highlight that the ability-residual framework can be combined with supervised ACD (with real affect labels) or contrastive CACD (without real affect labels), yielding response-prediction gains across all tested backbones in the reported main comparisons. On datasets with real affect labels, ARACD generally improves affect alignment, although not every backbone and metric benefits. Hierarchical ablation further suggests that the student-item residual layer is the main source of personalized response correction, while the concept residual layer complements concept-related bias. Furthermore, the experiments reveal that Q-matrix-constrained attention can aggregate relevant concept residuals in an interpretable way. Contamination leakage diagnostics reveal that removing ability residuals substantially increases the correlation and linear decodability between affective representations and synthetic cognitive bias. In contrast, the complete model transfers more of this stable bias to the ability residual pathway. PCA visualization, long-tail student reanalysis, and student-level cases further indicate that the framework not only improves overall prediction performance but also gives a more explicit separation of modeling roles in representation organization, low-interaction student diagnosis, and student-level correction.

Overall, this paper concludes that affective cognitive diagnosis should not focus only on improving response prediction but should also examine whether affective representations contain cognitive residuals that do not belong to affective states. Ability-residual decoupling provides a structured method for this problem, allowing the model to explain stable cognitive residual variation and model non-cognitive modulation, thereby improving the reliability, interpretability, and response prediction stability of affective diagnosis. Future work may further incorporate real multimodal affect observations and cross-platform educational data to evaluate the generalizability and limits of ability-residual decoupling in more complex learning scenarios.

\backmatter

\section*{Statements and Declarations}

\bmhead{Funding}
No funding was received for conducting this study.

\bmhead{Competing interests}
The authors have no relevant financial or non-financial interests to disclose.

\bmhead{Ethics approval and consent to participate}
Not applicable. This study used publicly available, de-identified educational datasets and involved no new data collection from human participants.

\bmhead{Consent for publication}
Not applicable.

\bmhead{Data availability}
The public datasets analysed in this study (ASSIST2017, ASSIST2012, ASSIST2009, and Junyi) are available from their original public repositories. The source code is available at \url{https://github.com/psychosiwa/ARACD}. Additional processed data and experimental outputs are available from the corresponding author on reasonable request.

\bmhead{Materials availability}
Not applicable.

\bmhead{Code availability}
The source code is publicly available at \url{https://github.com/psychosiwa/ARACD}.

\bmhead{Author contributions}
Boyuan Zhao: methodology, software, validation, formal analysis, investigation, data curation, visualization, and writing--original draft. Meng Ye: conceptualization, supervision, project administration, and writing--review and editing. Both authors read and approved the final manuscript.

\end{document}